\documentclass{article}

\usepackage[english]{babel}

\usepackage[letterpaper,top=2cm,bottom=2cm,left=3cm,right=3cm,marginparwidth=1.75cm]{geometry}

\usepackage{lineno}
\usepackage{siunitx}
\usepackage{enumitem}
\usepackage{graphicx}
\usepackage{amsmath, amssymb} % Add for \mathbb in formula
\usepackage{caption}
\usepackage{subcaption} % Add the subcaption package
\usepackage{authblk} % Author's affiliation
\usepackage{supertabular}
\usepackage{microtype}
\usepackage{hyperref}
\usepackage{float}
\usepackage{natbib} % Load natbib package with numerical citation option

\begin{document}

\title{Multi-label classification for multi-temporal, multi-spatial coral reef condition monitoring using vision foundation model with adapter learning}

% Authors
\author[1]{Xinlei Shao \thanks{These authors contributed equally to this work.}\thanks{8671299838@edu.k.u-tokyo.ac.jp}}
\author[*2]{Hongruixuan Chen}
\author[3]{Fan Zhao}
\author[4]{Kirsty Magson}
\author[5]{Jundong Chen}
\author[6]{Peiran Li}
\author[4]{Jiaqi Wang}
\author[1]{Jun Sasaki}

% Affiliations
\affil[1]{Department of Socio-Cultural Environmental Studies, Graduate School of Frontier Sciences, The University of Tokyo, Kashiwa, Chiba, 277-8563, Japan}
\affil[2]{Department of Complexity Science and Engineering, Graduate School of Frontier Sciences, The University of Tokyo, Kashiwa, Chiba, 277-8563, Japan}
\affil[3]{Department of Environment Systems, Graduate School of Frontier Sciences, The University of Tokyo, Kashiwa, Chiba, 277-8563, Japan}
\affil[4]{New Heaven Reef Conservation Program, Koh Tao, Surat Thani, 84360, Thailand}
\affil[5]{Data Science and AI Innovation Research Promotion Center, Shiga University, Hikone, Shiga, 522-8522, Japan.}
\affil[6]{Interfaculty Initiative in Information Studies, The University of Tokyo, Tokyo, 113-0033, Japan}

% Corresponding author note
\renewcommand\Authands{ and }
\renewcommand\Affilfont{\itshape\small}

\date{} % Removes the date
\maketitle

%\linenumbers

\begin{abstract}
\noindent Coral reef ecosystems provide essential ecosystem services, but face significant threats from climate change and human activities. Although advances in deep learning have enabled automatic classification of coral reef conditions, conventional deep models struggle to achieve high performance when processing complex underwater ecological images. Vision foundation models, known for their high accuracy and cross-domain generalizability, offer promising solutions. However, fine-tuning these models requires substantial computational resources and results in high carbon emissions. To address these challenges, adapter learning methods such as Low-Rank Adaptation (LoRA) have emerged as a solution. This study introduces an approach integrating the DINOv2 vision foundation model with the LoRA fine-tuning method. The approach leverages multi-temporal field images collected through underwater surveys at 15 dive sites at Koh Tao, Thailand, with all images labeled according to universal standards used in citizen science-based conservation programs. The experimental results demonstrate that the DINOv2-LoRA model achieved superior accuracy, with a match ratio of 64.77\%, compared to 60.34\% achieved by the best conventional model. Furthermore, incorporating LoRA reduced the trainable parameters from 1,100M to 5.91M. Transfer learning experiments conducted under different temporal and spatial settings highlight the exceptional generalizability of DINOv2-LoRA across different seasons and sites. This study is the first to explore the efficient adaptation of foundation models for multi-label classification of coral reef conditions under multi-temporal and multi-spatial settings. The proposed method advances the classification of coral reef conditions and provides a tool for monitoring, conserving, and managing coral reef ecosystems.

\end{abstract}

\section{Introduction}
Despite occupying less than 1\% of the ocean floor, coral reef ecosystems are the most biodiverse marine habitats and provide human beings with ecological, economic, and cultural values \citep{Cesar2004importance, Laurans2013services, Seenprachawong2016economic}. Coral reef ecosystems provide habitats for many marine species, such as crustaceans, mollusks, echinoderms, and fish. These species serve as food sources for coastal populations and act as key species to support the resilience and resistance of marine ecosystems \citep{White2000threat}. Coral reefs are also considered green infrastructure, protecting coastal regions from extreme weather such as storm surges and hurricanes \citep{Brathwaite2022infra}. Despite their importance, the health condition of coral reef ecosystems is degrading due to climate change and human activities \citep{Wilkinson1996impact}. To address these issues, coral reef conservation and restoration activities are ongoing worldwide, with community-based programs proven to be one of the most effective conservation methods \citep{Baums2008community}. The citizen science-based program is a community-based coral reef conservation activity where scientists and trained volunteers (i.e., citizens) work together to collect and analyze data \citep{Hesley2023citizen}. Well-known examples of such programs include Reef Check, REEF, Conservation Diver, BleachWatch, and CoralWatch, which are implemented worldwide \citep{Hesley2017coralwatch, Braverman2022citizen}. Volunteers receive training in marine biology and conservation practices in these programs before conducting regular underwater surveys to monitor coral reef conditions \citep{Vieira2020citizen}. These monitoring activities are crucial, as they provide data on changes in coral coverage, abundance, and diversity, which form the basis for subsequent conservation actions \citep{Reshma2023importance}. Despite the variety of programs, their monitoring protocols and assessment standards are generally consistent. Most coral reef monitoring efforts use the underwater transect survey method, where volunteer divers document coral reef conditions through text, photos, or videos \citep{Rogers1983transect, ramos2010transect, Forrester2015transect}. However, this approach has significant limitations, including high labor demands, low classification efficiency, and inconsistencies in data quality. These challenges arise because the entire process is highly dependent on the expertise and subjective judgment of human experts \citep{Done2017dataquality, FiguerolaFerrando2024dataquality}. Addressing these challenges through automated, accurate, and effective image classification based on universal assessment standards has therefore become a research priority for the monitoring and conservation of coral reef ecosystems.

Traditional methods for automatic image classification primarily relied on manual feature extraction and classical machine learning techniques such as support vector machines (SVMs), k-nearest neighbors (KNNs), decision trees (DTs), random forest (RF), and artificial neural networks \citep{Maxwell2018machinelearning, Sheykhmousa2020machinelearning, Goel2023machinelearning}. Although these traditional machine learning methods achieved good performance in structured, smaller datasets, the emergence of deep learning-based models further improved classification performance. This improvement is attributed to their scalability, the ability to learn sophisticated and hierarchical features directly from input data, and the elimination of manual feature engineering \citep{Houssein2021DL}. Specifically, through the supervised learning process of the deep learning model, it can automatically extract and learn features from the input data; once trained, these models can classify images into predefined labels without relying on domain experts \citep{Lv2022DLreview}. Deep learning models have been applied in a wide range of fields to address practical real-world needs, such as smart agriculture \citep{Dhanya2022agricultre}, medical image analysis \citep{meyer2014medical}, autonomous driving \citep{Grigorescu2020driving}, natural language processing \citep{Otter2021NLP}, and ecological monitoring \citep{zhao2024ecology}. Among these, the classification of coral reef conditions is an example within the field of ecological monitoring. 

Although conventional deep learning models show great potential and have been applied to various ecological monitoring tasks \citep{Palanisamy2021ecology}, such as image classification, object detection, and semantic segmentation, significant challenges remain in achieving high accuracy. These challenges primarily arise from the complexity of underwater images \citep{González-Sabbagh2023underwater, Naveen2024underwater}, the high intra-class and inter-site variability in coral morphology \citep{Shihavuddin2013variability}, the sophisticated features required to classify coral reef images \citep{Gómez-Ríos2019coral}, and the difficulty of adapting trained deep models to different seasons or regions \citep{Long2015transfer, Singhal2023transfer}. The emergence of foundation models offers promise in addressing these challenges that conventional deep learning models face. Foundation models are large models pre-trained on large amounts of general domain data to gain high generalizability across different downstream tasks and transferability to various domains \citep{bommasani2021foundation, Koubaa2023chatgpt}. These models comprise numerous trainable parameters that can be fine-tuned using a specific training dataset to be adapted to the corresponding domain \citep{Englert2024generalizability}. One of the well-known examples is ChatGPT, which is a large language model (LLM) built upon the transformer architecture and pre-trained on diverse and extensive publicly available text data; it shows great generalizability across unseen tasks such as translation, coding, story writing, and document processing \citep{wu2023chatgpt}. Similarly, with advances in computer vision, various vision foundation models have been proposed, such as the Segment Anything Model (SAM) \citep{kirillov2023SAM}, Mask Autoencoders (MAE) \citep{he2022MAE}, DINO \citep{Caron2021DINO}, and DINOv2 \citep{oquab2024dinov2}. DINO is a self-supervised learning method within these foundation models to train the Vision Transformer (ViT) \citep{dosovitskiy2021ViT} with knowledge distillation without labels. It is characterized by its Transformer-based architecture and a teacher-student network design. The student network continuously updates its parameters through an exponential moving average to match the output of the teacher network. This design enables DINO to learn rich visual information from images without labeled data, achieving state-of-the-art (SOTA) performance in various application fields such as computational pathology in the medical field \citep{chen2024pathology, rabbani2024surgical}, leaf disease identification \citep{Chen2023foundation}, and plant recognition in the agricultural field \citep{Ye2024plantvision}, and defect detection in the manufacturing industry \citep{Pasanisi2024manufacturing}. DINOv2 is an improved framework based on DINO, improving data curation, training efficiency, domain adaptation, and model stability. Compared to the original DINO, DINOv2 demonstrates significant potential to achieve high performance in the classification task of coral reef conditions. This is achieved by fine-tuning the model with domain-specific data, such as labeled coral reef condition data.

Despite the high accuracy and generalizability of vision foundation models like DINOv2, these models can contain up to billions of trainable parameters \citep{tsouvalas2024parameters}. As a result, fine-tuning such models requires significant computational resources, and the associated carbon emissions during this process are substantial. For example, training a SOTA foundation model named SatMAE takes 768 hours on a V100 GPU, with a total estimated carbon footprint of 109.44 kg CO$_2$ \citep{mendieta2023carbonemission}. To make vision foundation models more accessible to resource-limited users, such as conservation communities or reef managers, and to reduce carbon emissions during training, an adapter-based parameter-efficient fine-tuning (PEFT) method called Low-Rank Adaptation (LoRA) \citep{hu2021lora} was introduced. LoRA updates only a small subset of parameters compared to the total number of trainable parameters in vision foundation models, enabling these models to adapt to domain-specific tasks with relatively low computational resource requirements. Considering coral reef conservation communities and government sectors as end users, this study applied LoRA to fine-tune DINOv2, thereby reducing the use of computational resources and carbon emissions during the training process.

The main contributions of this study are as follows.
\begin{itemize}
    \item Dataset: We establish an open-source, multi-temporal coral image dataset based on underwater photo transect surveys conducted in Koh Tao, Thailand, in the Indo-Pacific region. The images, captured over five months to represent dry and wet seasons, were annotated by domain experts following coral reef conservation standards \citep{scott2019ecological, haskin2022coral}.
    \item Vision foundation model: This study is the first to explore the potential of adapting the vision foundation model (DINOv2) to coral reef condition classification tasks, providing a benchmark in this field. Meanwhile, comparing conventional deep models and vision foundation models provides insight into the trade-offs between performance and computational costs. These findings aim to guide decision-makers in choosing between task-specific and highly generalizable foundation models.
    \item Model fine-tuning: We investigates using LoRA for fine-tuning DINOv2 on this multi-label classification task. The results demonstrate the effectiveness of LoRA in significantly reducing computational costs. 
    \item Tool for ecological monitoring: By open source the data set and code, this study provides an accessible tool for automating the classification of coral reef conditions and stressors. The proposed method aligns with universal standards for coral reef conservation, enhancing its usability for ecological monitoring and decision-making.
\end{itemize}

\section{Related work}
\subsection{Machine learning models}
Traditional methods for classifying coral reef conditions rely on visual inspection and manual classification of images conducted by domain experts. Due to the development of machine learning, researchers have started automating the classification process. \citet{Marcos05ML} first proposed a feedforward backpropagation neural network to classify underwater images as living coral, dead coral, and sand. They achieved a success rate of 86.5\% while a false positive is 6.7\%). \citet{stough2012machinelearning} proposed a coral segmentation model based on linear support vector machines, using quantile functions and SIFT texture features as feature extraction methods. Their model achieved an overall average precision (AP) of 72.1\% in segmenting \textit{Acropora cervicornis} from the background. \citet{Tusa2014machinelearning} also focused on distinguishing coral from non-coral features. They tested nine popular machine-learning algorithms for coral detection and found that the decision trees achieved the best balance between accuracy and computational costs. Similarly, \citet{Beijbom2012ML} and \citet{Wahidin2015ML} have done similar work. However, these studies focus on the classification of benthic composition, while few studies address the classification of coral reef conditions using machine learning. This is likely due to the inherent complexity of this task, such as the limited feature extraction capabilities of traditional machine learning methods, the complexity and heterogeneity of coral reef images, and the lack of high-quality underwater image datasets \citep{Chegoonian2017challenge}.

\subsection{Conventional deep learning models}
Deep learning models have significantly improved performance over traditional machine learning methods, demonstrating great potential in automating coral reef-related classification tasks, such as identifying coral species, fish species, and benthic composition. Convolutional Neural Networks (CNNs) and Transformer-based models have been widely used in this domain \citep{Mahmood2017coralclassification, Arsad2023classification, shao2024coral}. Existing studies can be summarized mainly in two aspects: (1) quantitative analysis by segmenting the benthic composition \citep{Williams2019bentho, Wang2023segmentation, Sauder2024segment, Šiaulys2024bentho} or coral genera \citep{Raphael2020genera, Reshma2023importance, Zhang2024segmentation} to derive distribution maps, and (2) qualitative analysis by detecting compromised coral reefs through various deep learning models \citep{Fawad2023bleached, shao2024coral}. 

From the perspective of segmentation of benthic composition and coral genera, most studies focused on creating benthic habitat maps or coral taxonomy. \citet{Schürholz2023betho} proposed a workflow for benthic habitat mapping through two machine learning methods, which are a segmented method based on ensemble learning and a patched method built on a spectral-spatial residual network \citep{Zhong2018pachted}. The workflow presented in their study successfully segmented the underwater hyperspectral images into 43 classes, including coral from various taxonomic families, genera, or species, and substrates such as sediment, algae, sponges, cyanobacterial mats, or turf algae. Similarly, \citet{Pierce2021bentho} presented a method based on fully convolutional networks (FCNs) for the semantic segmentation of 3D coral reef habitat models; this method enables accurate segmentation of 3D models into five classes, including both coral species and functional groups. There are also many studies on classifying the coral genus or species using deep learning models \citep{Mahmood2017species, Gómez-Ríos2019species, Raphael2020species, Arsad2023species}, where different deep learning networks were tested and applied to classify underwater images into several classes of coral genus or species. Specifically, \citet{Li2024transformer} evaluated the performance of three Transformer-based models, together with a transfer learning technique to segment coral species and growth forms. Despite the overfitting issues found in these models, their study established a benchmark for assessing the performance of Transformer architectures in coral species identification tasks. Although these studies provide valuable information on coral coverage, community composition, species richness, and community structure \citep{Palma2017bentho}, there are limitations in providing targeted information specifically designed for coral reef conservation. The interests of citizen science-based coral reef conservation programs lie in monitoring the coral reef conditions and corresponding stressors, which reflect the health status of the habitats and are essential to guide subsequent conservation activities \citep{Tun2005monitoring, scott2019ecological}. For example, if compromised coral reefs and predators are frequently detected in images taken at certain sites, conservationists can visit the site and remove predators. 

Despite a limited number of studies, some conducted qualitative analysis by identifying compromised coral reefs and stressors using object detection or image classification models. \citet{Mary2018disease, Mary2019disease} constructed two efficient feature descriptors and input the extracted features into 11 different traditional machine learning classifiers to test their performance in classifying diseased coral reef images into one of nine disease classes. Consequently, the pulse-coupled convolutional neural network (PCCNN) released the highest overall accuracy of 92\%. While PCCNN belongs to the conventional machine learning technique, advances in deep learning models have also demonstrated good performance in classifying coral reef conditions. \citet{Borbon2021coralhealth} constructed a self-designed CNN architecture to classify coral reef images into healthy, bleached, and dead classes. They tested their model on two open-source datasets containing 150 and 185 images and achieved accuracy of 84.93\% and 68.75\%, respectively. \citep{Kaushik2024VGG19} developed a VGG19-based model to classify 923 images into two classes, which are healthy or bleached, and their model achieved an accuracy of 74\% while showing potential in real-time monitoring capabilities. Similarly, an InceptionV3-based model was implemented by \citet{Kaur2024inceptionv3} to classify 970 coral reef images into healthy or bleached, and their model achieved a promising accuracy of 97.3\%. Although these studies provided valuable information on coral reef health classification, limitations exist as they focused only on two to three classes, which is inadequate to provide a comprehensive understanding of coral reef conditions for conservationists. Meanwhile, these are single-label classification tasks in which each image is assigned to a single class. In contrast, monitoring the condition of coral reefs often requires identifying multiple coexisting health conditions and stressors within an image \citep{scott2019ecological}, which is regarded as a multi-label classification task.

Multi-label classification has been applied in various fields, such as microbiome classification \citep{wu2021multimicrobiome}, species identification through sounds \citep{xie2017multifrog, Swaminathan2024multibird}, species identification through images \citep{zhang2016multibird, Alfatemi2024multibird, Xu2024multibenthic}. Specifically, \citet{Xu2024multibenthic} applied several self-supervised learning (SSL) methods, including SimSiam, BYOL, and MoCo-v3, to pre-train ResNet-50 and Vision Transformer-Base (ViT-B) models on the BenthicNet dataset for hierarchical multi-label classification of benthic imagery of coral reef ecosystems. The ViT-B model with MoCo-v3 achieved the best performance with an average accuracy of 85.5\% and a macro F1 score of 68.7\%. \citet{shao2024coral} evaluated the performance of several SOTA deep learning architectures for multi-label classification of coral reef conditions and found that Swin-Transformer-Base achieved the best performance. Further performance improvement was achieved by an ensemble model that integrates Swin-S, Swin-B, and EfficientNet-B7, resulting in a match ratio of 63.33\% and a macro F1 score of 81.74\%. These studies achieved promising accuracy and provided benchmarks for guiding the development of multi-label classification models in the relevant fields. However, these proposed models are built on conventional deep learning architectures.  

\subsection{Vision foundation models}
The concept of the vision foundation model was first widely recognized in 2021 due to the release of Contrastive Language–Image Pre-training (CLIP) by OpenAI \citep{openai2021clip}. CLIP is a milestone marking the extension of the application of the foundation model from natural language processing to computer vision tasks. Since then, many studies have been conducted to adapt vision foundation models to a variety of downstream tasks. \citet{Chen2023foundation} tested the performance of three vision foundation models, MAE, DINO, and DINOv2, in leaf counting, instance segmentation, and disease classification tasks. They found that DINOv2 with LoRA achieved the best performance in leaf counting and disease classification, which achieved a percent accuracy of 47.6\% and a general precision of 90.3\%, respectively. \citet{doherty2024foundation} tested several vision foundation models on a drone-collected leafy spurge dataset, and they found that fine-tuned DINOv2 achieved an accuracy of 85\%, while GPT-4o achieved 75\% accuracy with only eight examples per class. Their work demonstrated the potential for applying the vision foundation model for ecological monitoring in data-limited scenarios. Although their work showed superior performance of the vision foundation model in single-label classification, the performance in multi-label classification remains unknown. Furthermore, multi-temporal and multi-spatial monitoring of coral reef conditions is an Out-of-Distribution (OOD) task \citep{Borlino2024OOD}, where images are collected under varying lighting, angles, and geographical environments, with previously unseen objects appearing across different seasons or habitats (e.g., different competitors in the wet season compared to the dry season). Therefore, this study adopted the vision foundation model for the multi-label classification of coral reef conditions.

\section{Methodology}
\begin{figure}[ht]
\centering
\includegraphics[width=1\textwidth]{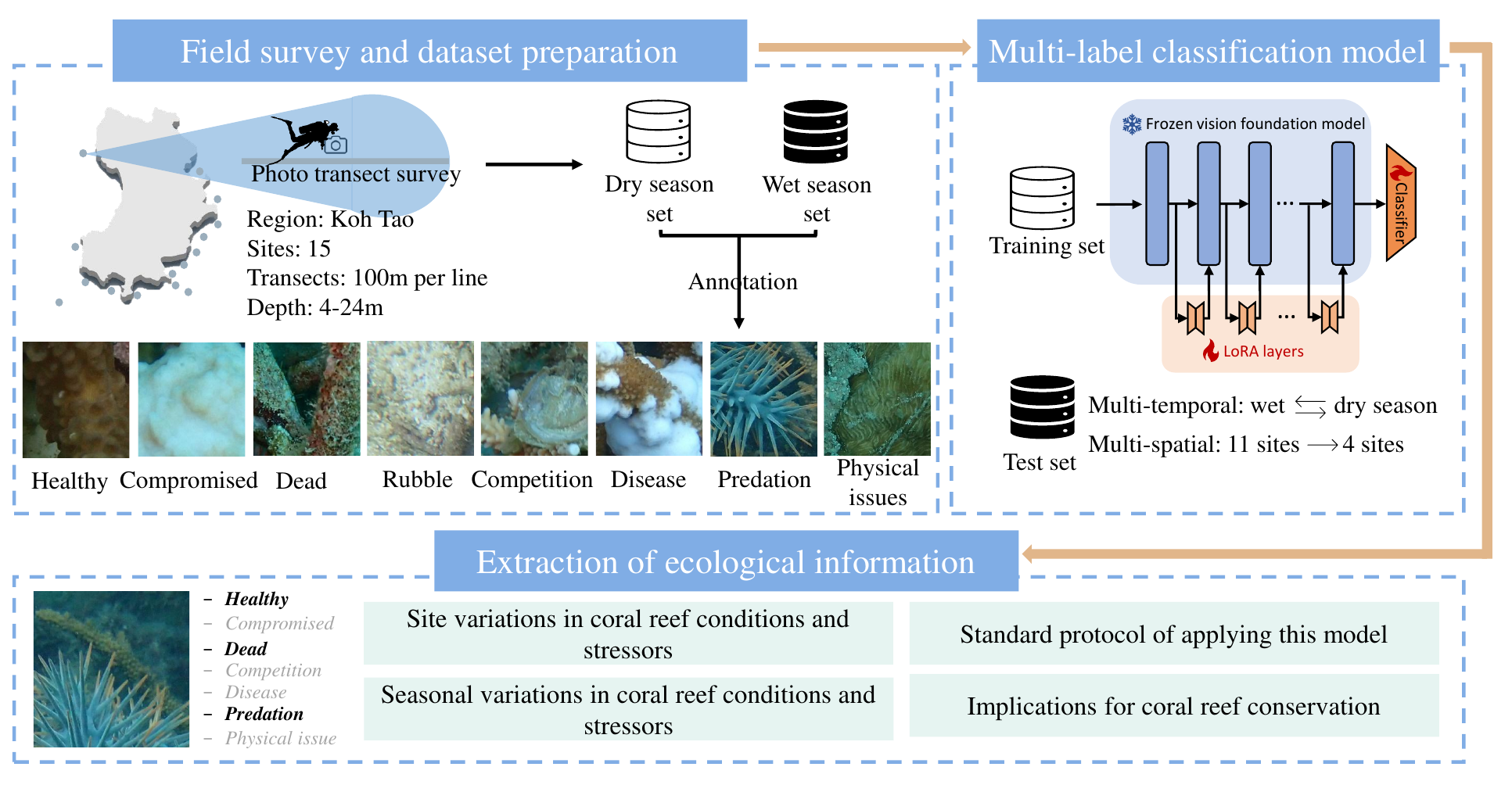} 
\caption{Flowchart illustrating the process of establishing the proposed method in this study.}
\label{fig:flowchart} 
\end{figure}

The workflow of this study is illustrated in Figure \ref{fig:flowchart}, and consists of three parts, including the field survey to collect multi-temporal and multi-spatial underwater images for model training, the establishment of multi-label classification models, and the extraction of ecological information from the proposed models.

\subsection{Study site and dataset}
\begin{figure}[ht]
\centering
\includegraphics[width=1\textwidth]{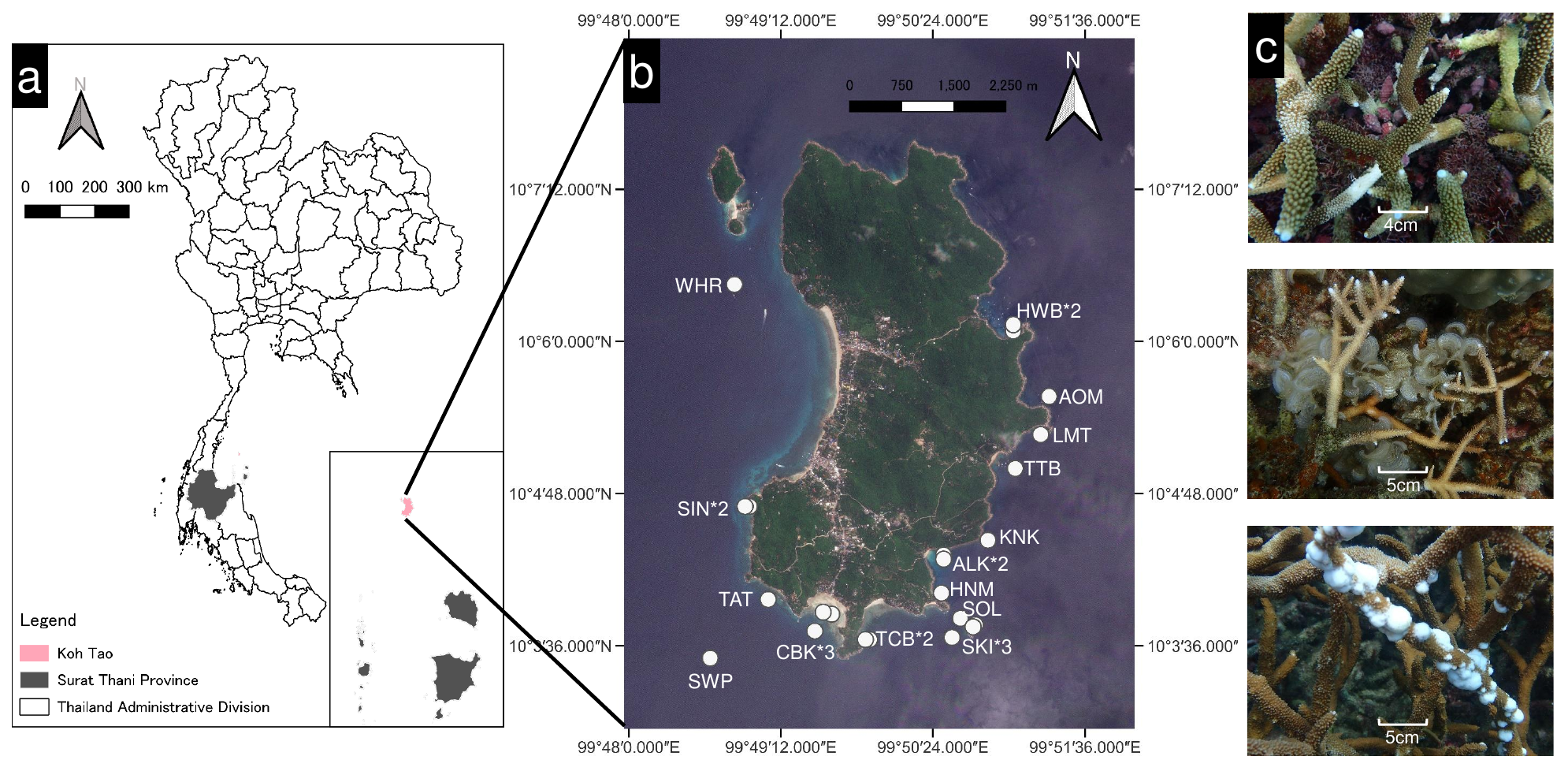} 
\caption{(a) Location of the study area, Koh Tao, Thailand; (b) Surveyed dive sites at Koh Tao (base map sourced from \citet{PlanetLabs2024}); (c) Typical stressed coral reef images taken through underwater photo transect method.}
\label{fig:study_area} 
\end{figure}

The dataset used in this study is based on fieldwork conducted in Koh Tao, where underwater images were collected to build the dataset to establish our deep learning models. As shown in Figure \ref{fig:study_area}, Koh Tao is an island located in the western area of the Gulf of Thailand, Surat Thani province, Thailand. It is a 21 km\(^2\) island, surrounded by fringing reefs with high biodiversity, making it a popular diving destination \citep{Tapsuwan2015tourism, Mehrotra2021kohtao}. The climate of Koh Tao is tropical monsoon, which is characterized by warm temperatures ($25^{\circ}\text{C}$ to $35^{\circ}\text{C}$ ) and two main seasons. The wet season generally begins at the end of October and lasts until the end of January, while the dry season takes the remaining months throughout the year \citep{Buranapratheprat2008climate}. During the wet season of the island, the southwest monsoon brought strong precipitation, leading to increased terrestrial runoff, algae bloom, and low water clarity \citep{Sojisuporn2010currents, Leenawarat2022monsoon}. During the dry season, the island is affected by the northeast monsoon, which results in low precipitation, calm waters, and high water clarity, thus enhancing the quality of the underwater images collected.

\begin{figure}[ht]
\centering
\includegraphics[width=0.7\textwidth]{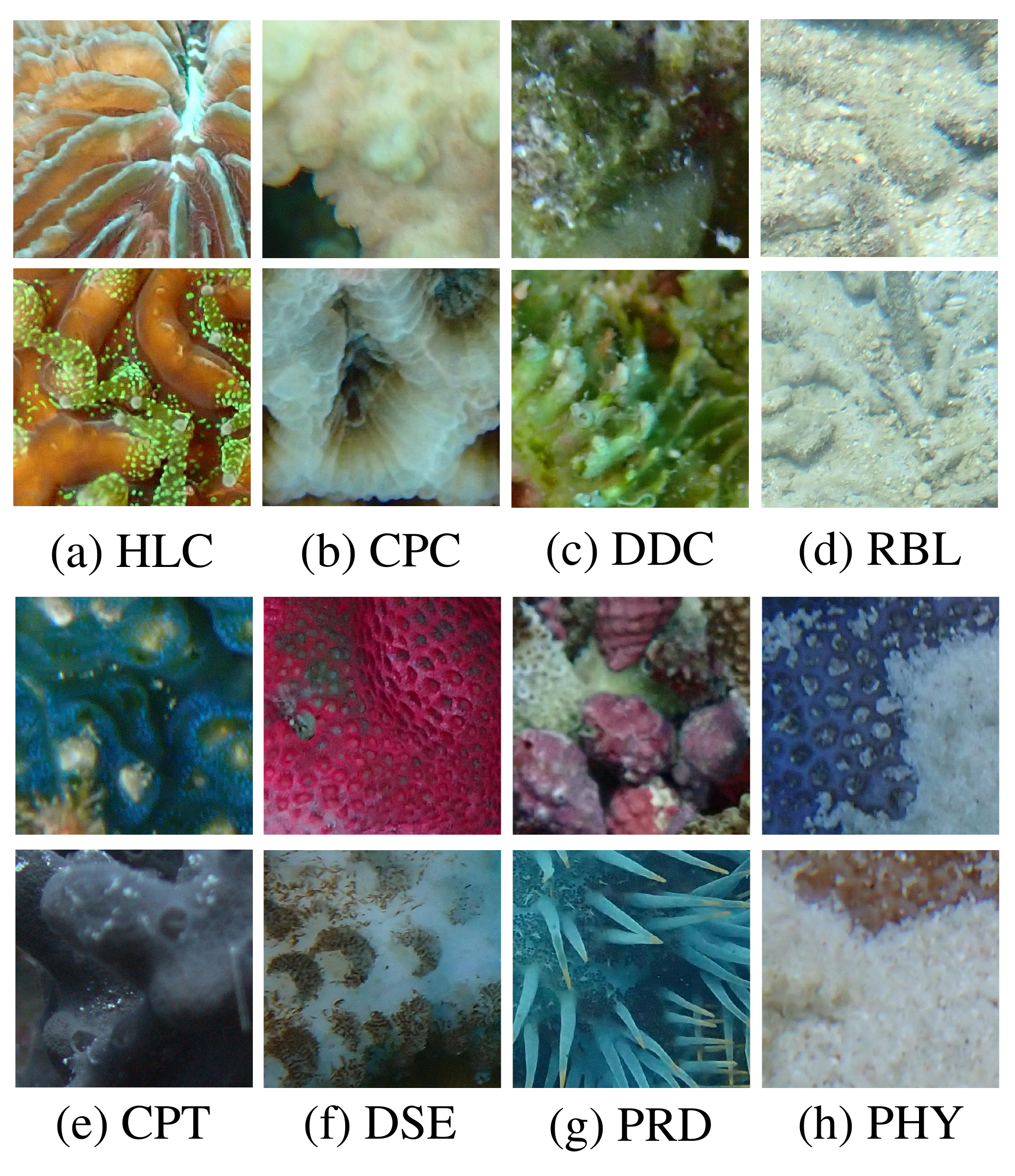} 
\caption{Sample image patches from the dataset, each column displaying two examples per class. (a) Healthy Coral (HLC), (b) Compromised Coral (CPC), (c) Dead Coral (DDC), (d) Rubble (RBL), (e) Competition (CPT), (f) Disease (DSE), (g) Predation (PRD), and (h) Physical Issues (PHY). As this is a multi-label classification task, each image can contain more than one class; for example, the class ``compromised coral" often coexists with the class ``disease."}
\label{fig:sample_patches} 
\end{figure}

Several dive sites were surveyed at different times to examine spatial and temporal variations in coral reef conditions and dominant stressors. The survey sites covered 15 dive sites in Koh Tao (reflected in Figure \ref{fig:study_area}), and the water depth ranged from 3.6 m to 23.3 m, with an average water depth of 12.1 m. Underwater surveys were conducted over three periods of time, April 2023, August to September 2023, and January to February 2024. The survey method adopted in this study is the underwater photo-transect survey method, which is a widely applied sampling method that sends divers to monitor the coral reefs following a transect line. Specifically, this study followed the same protocol applied by coral reef conservation programs \citep{scott2019ecological}, where a $100 \text{m length} \times 5 \text{m width}$ section was examined in each survey. The images were taken by divers with an Olympus Tough TG-6 underwater camera with manual white balance whenever the water depth was changed. In this study a total of 23 surveys were conducted, producing 1,203 images with a resolution of $4000 \times 3000$ pixels. Each image was cut into 35 patches, resulting in 42,105 patches with a resolution of $512 \times 512$ pixels each. Specifically, the ratio of dry season images to wet season images is 3 to 1, which aligns with the respective durations of these seasons (nine months for the dry season and three months for the wet season). There are eight classes involved in this multi-label classification task. These are divided into two groups: four classes that represent the status of coral reefs, which are ``Healthy coral," ``Compromised coral," ``Dead coral," and ``Rubble", and four classes that represent stressors affecting coral reef conditions, which are ``Competition," ``Disease", ``Predation," and ``Physical issues." Figure \ref{fig:sample_patches} shows some sample image patches from this dataset.

\subsection{Proposed multi-label classification method}
\subsubsection{DINOv2}
DINOv2 is one of the recently proposed foundation models introduced by \citet{oquab2024dinov2}. It is a groundbreaking foundation model in computer vision, aimed at generating versatile visual features that perform robustly across both image- and pixel-level tasks without requiring task-specific adaptations. The authors devised an innovative data curation pipeline to construct a diverse and high-quality dataset, enabling effective self-supervised learning. ViT model collections were trained to optimize feature learning using an efficient and scalable self-supervised methodology. The DINOv2 model family, including distilled smaller variants, demonstrated superior performance across numerous benchmarks, setting a new standard for general-purpose visual representations in the field. Although foundation models such as DINOv2 have been applied for classification tasks and have demonstrated superior performance in various research domains, their applications are still limited in ecological monitoring. Therefore, this study first attempts to develop a novel model based on DINOv2 to classify the condition of the coral reef. Specifically, this study used the DINOv2-Giant architecture, and all subsequent references to DINOv2 in the context of this study specifically denote the DINOv2-Giant variant.

\subsubsection{LoRA}
LoRA was first proposed by \citet{hu2021lora} and applied to fine-tune LLMs by introducing trainable rank decomposition matrices into each layer of the Transformer architecture for fine-tuning, enabling adaptation of the model to specific tasks while keeping the original pre-trained weights frozen \citep{sun2024lora}. Specifically, instead of directly updating the original pre-trained weight matrix $W_0\in R^{d \times k}$, LoRA adds a trainable low-rank decomposition matrix $\Delta W=B \cdot A$ to adapt to the downstream tasks, where $B \in R^{d \times r}$, $A \in R^{r \times k}$, and the low-rank $r\ll min(d,k)$. During the training stage, $W_0$ remains frozen and unmodified, which ensures that the general knowledge of the large model learned during the pre-training is fully retained. Meanwhile, the low-rank matrices $A$ and $B$ are trained and updated. Therefore, for an input $x$, in a forward pass modified by LoRA, the output vector $h$ is described as:

\begin{equation}
\label{eq:lora_h}
h = W_0 x + \Delta Wx = W_0 x + BAx
\end{equation}

In this case, adapting LoRA can reduce trainable parameters from the full weight matrix $d \times k$ to $(d \times r + r \times k)$. Since the low-rank number $r$ is much smaller than $d$ and $k$, the number of trainable parameters is drastically reduced, significantly reducing computational costs during the fine-tuning process.

\subsubsection{Network architecture}
\begin{figure}[ht]
\centering
\includegraphics[width=1\textwidth]{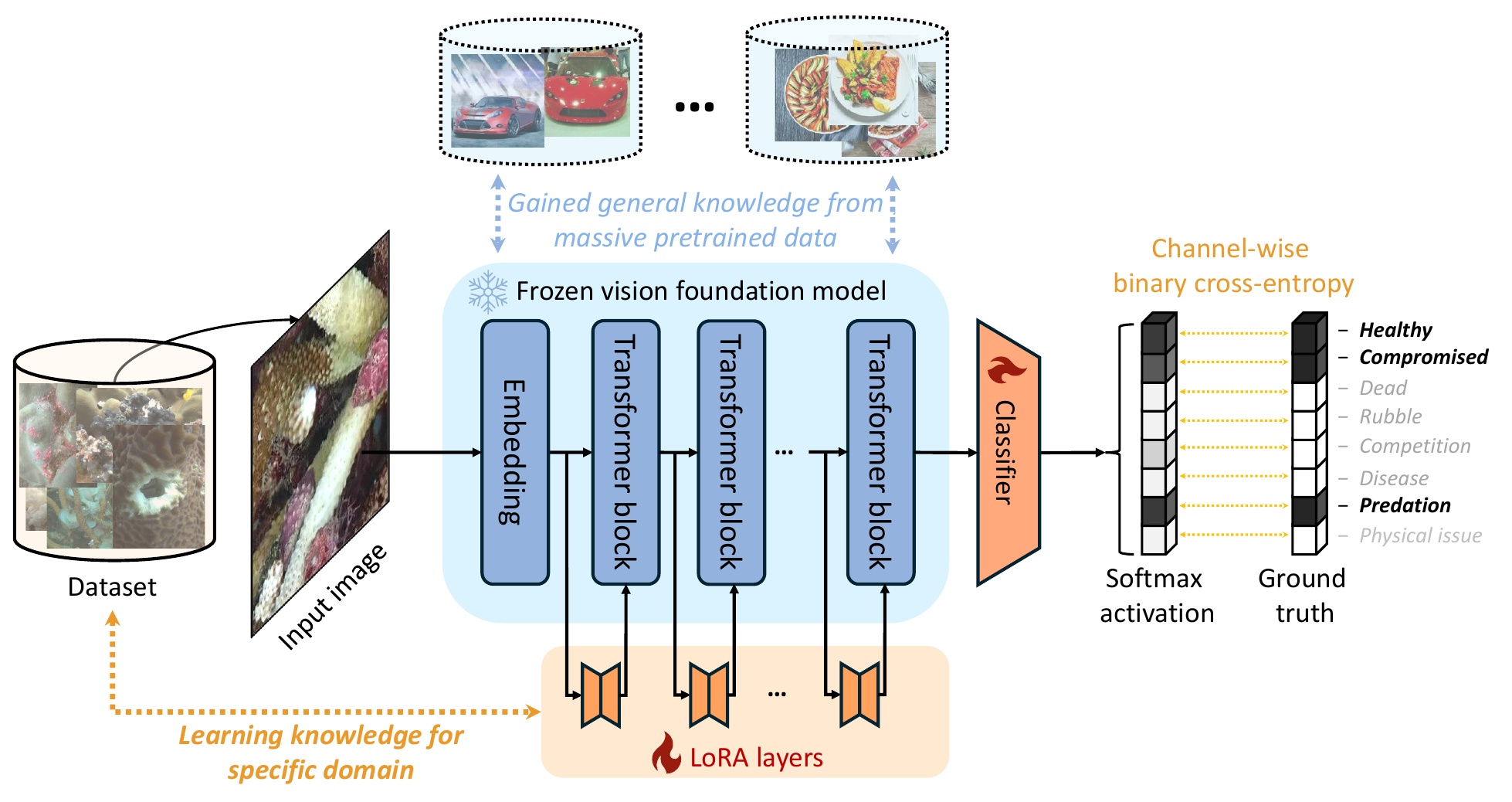} 
\caption{Framework illustrating the process of developing the foundation model for the multi-label classification task. During the fine-tuning process, the pre-trained weights of ViT are frozen, while only LoRA layers and classifier are trained. With the adaptation of LoRA, we can realize the goal of exploiting the general knowledge already available in the foundation model, while making it adaptable to our task at a very low training cost.}
\label{fig:proposed_method} 
\end{figure}

The network architecture of our proposed model is illustrated in Figure \ref{fig:proposed_method}. Given a coral reef image $x \in \mathbb{R}^{H \times W \times C}$ as input, where its spatial resolution is $H \times W$ and channels are $C$, the model aims to predict a set of labels $\hat{y} \in \{0, 1\}^C$, where $C$ is the number of classes, ensuring the predictions align closely with the ground truth labels. DINOv2 is selected as the image encoder, where the input images are divided into patches of size $512 \times 512$ pixels and then flattened and passed through a linear projection. Positional embeddings are added to the tokenized representations, and a learnable class token is used to capture the global image context. These token embeddings are processed through a series of Transformer blocks, generating feature representations optimized for predicting multiple labels simultaneously. During training, the pre-trained weights of DINOv2 are frozen, while the weights of LoRA layers, added to each Transformer block, are updated to capture features and semantic information. The LoRA layers significantly reduce the total number of trainable parameters in the architecture by compressing the Transformer features into a low-rank space and then re-projecting them to match the output feature dimensions of the frozen Transformer blocks. A classification decoder is appended at the end to assign multi-labels to the input images.

\subsubsection{Loss function}
In this study, the binary cross-entropy loss function is used to train all multi-label classification models. The equation for the loss function is given as:

\begin{equation}
\label{eq:bicroentropy_loss_function}
\mathcal{L} = -\frac{1}{N} \sum_{i=1}^N \sum_{j=1}^C \left[ y_{ij} \log(\hat{y}_{ij}) + (1 - y_{ij}) \log(1 - \hat{y}_{ij}) \right]
\end{equation}
where $N$ denotes the number of samples in the dataset, $C$ denotes the total number of classes, $y_{ij}$ denotes the ground truth label for the $j-th$ class of the $i-th$ sample ($y_{ij}\in {0,1}$, indicating whether the class is relevant), $\hat{y}_{ij}$ denotes the predicted probability of the $j-th$ class for the $i-th$ sample (output of a sigmoid function, $y^{ij}\in[0,1]$). Although multiple classes can coexist within a single sample, the binary cross-entropy function assumes that each class (or channel) is independent. Consequently, each class can be treated as a separate binary classification problem, making the sigmoid activation function a suitable choice for the output layer.

\subsection{Conventional deep learning models as baseline}
Several conventional deep learning models were established as baselines to assess and compare the performance of different models in the multi-label classification of coral reef conditions. These models evaluate the performance enhancement of the DINOv2-LoRA model, which is based on the vision foundation model relative to conventional task-specific deep learning models. These baseline models include renowned architectures such as VGG \citep{simonyan2015VGG}, Inception V3 \citep{szegedy2016Inceptionv3}, ResNet \citep{he2016ResNet}, DenseNet \citep{huang2017densenet}, EfficientNet \citep{tan2019efficientnet}, HRNet \citep{Wang2021HRNet}, and Swin-Transformer \citep{liu2021swintransformer}. These architectures cover most relevant studies and include CNN and Transformer-based structures.

\subsection{Experimental setup}
For the experiment using the all-season mixup dataset, the dataset was divided into training, validation, and test sets in a 7: 1: 2 ratio. For experiments testing transfer learning capability, the dataset was split differently, with detailed information provided in Section \ref{sec:multi-temporal} and Section \ref{sec:multi-spatial}. All conventional deep learning models were initialized with pre-trained weights from ImageNet, while DINOv2 was pre-trained on the LVD-142M dataset \citep{oquab2024dinov2}. Subsequently, all models were fine-tuned using PyTorch. The binary cross-entropy loss function was applied to each unit in the network output as a loss function. AdamW Optimizer \citep{kingma2014adamoptimizer} was used to train all deep learning models, with a learning rate of $5\times 10^{-6}$ for DINOv2 and $1\times 10^{-4}$ for the other models. The training setup included a weight decay of $5\times 10^{-4}$, a batch size of 16, and 30,000 training iterations. All experiments were conducted on an NVIDIA A100 Tensor Core GPU, boasting 504 GB of RAM and powered by 80 CPU processors (Intel(R) Xeon(R) CPU E5–2698 v4 @ 2.20GHz).

\subsection{Evaluation metrics}
Three commonly used evaluation metrics for multi-label classification, which are match ratio, Macro F1 score, and Micro F1 score, are applied to assess and compare the performance of various established deep learning models \citep{naidu2023evaluation}. Match ratio measures whether the set of predicted labels by the model exactly matches the set of ground truth labels for each instance. It is given by:
\begin{equation}
\label{eq:match_ratio}
\text{Match Ratio} = \frac{\text{Number of Correct Predictions}}{\text{Total Number of Predictions}}
\end{equation}

Both the Macro F1 score and the Micro F1 score are derived from precision and recall, which are core measures of a model's performance. The difference lies in the way they treat each class. Specifically, the Macro F1 score calculates the F1 score independently for each label and then averages them, whereas the Micro F1 score aggregates precision and recall of all classes to derive a single F1 score. Calculating the Macro F1 score involves two steps: computing the F1 score independently for each class $i$ as described in Equation \ref{eq:precision_macro}, \ref{eq:recall_macro} and \ref{eq:F1_macro}; calculating the arithmetic mean of these scores as shown in Equation \ref{eq:macro F1}. Assuming TP = True Positive, FP = False Positive, FN = False Negative, $N$ = the total number of classes, and $i$ is the index of each class in the dataset, the formulas are given as follows:
\begin{equation}
\label{eq:precision_macro}
\text{Precision}_i = \frac{\text{TP}_i}{\text{TP}_i + \text{FP}_i}
\end{equation}

\begin{equation}
\label{eq:recall_macro}
\text{Recall}_i = \frac{\text{TP}_i}{\text{TP}_i + \text{FN}_i}
\end{equation}

\begin{equation}
\label{eq:F1_macro}
\text{F1}_i = 2\times \frac{{\text{Precision}_i \times \text{Recall}_i}}{{\text{Precision}_i + \text{Recall}_i}}
\end{equation}

\begin{equation}
\label{eq:macro F1}
\text{Macro F1} = \frac{1}{N} \sum_{i=1}^{N} \text{F1}_i
\end{equation}

Calculating the Micro F1 score involves two steps: calculating Precision and Recall across all classes as given by Equations \ref{eq:precision_micro} and \ref{eq:recall_micro}; determining the Micro F1 score based on global Precision and Recall derived in the previous step.

\begin{equation}
\label{eq:precision_micro}
\text{Precision} = \frac{\sum_{i=1}^{N} \text{TP}_{i}}{\sum_{i=1}^{N} \text{TP}_{i} + \sum_{i=1}^{N} \text{FP}_{i}}
\end{equation}

\begin{equation}
\label{eq:recall_micro}
\text{Recall} = \frac{\sum_{i=1}^{N} \text{TP}_{i}}{\sum_{i=1}^{N} \text{TP}_{i} + \sum_{i=1}^{N} \text{FN}_{i}}
\end{equation}

\begin{equation}
\label{eq:micro F1}
\text{ Micro F1} = 2\times \frac{\text{Precision} \times \text{Recall}}{\text{Precision} + \text{Recall}}
\end{equation}

\section{Results}
In this study, three experimental settings were established to compare and evaluate the performance and transfer learning capability of our proposed model. These settings differ in their datasets as follows: (1) a mixup dataset containing all-season, all-site data to compare the overall performance of different models; (2) separate datasets for the dry season and winter season to assess the transfer learning ability of different models on multi-temporal data; (3) habitat-specific datasets to evaluate the transfer learning ability of different models on multi-spatial data. Accordingly, the results section is organized in the same sequence as described above.

\subsection{Comparison of performance in baseline models and DINOv2-LoRA}
The mixup dataset, containing all field data regardless of collection time or location, was used to evaluate the overall performance of different models on the coral reef condition multi-label classification task. According to the results shown in Table \ref{tab:mix_performance}, the vision foundation model (DINOv2 and DINOv2-LoRA) outperformed all other conventional deep learning models. In general, DINOv2-LoRA achieved the best performance in match ratio, micro F1, and macro F1 scores, as well as for most classes, except for achieving F1 scores that were 0.01\% and 0.56\% lower than DINOv2 in the ``HLC" and ``CPT" classes, respectively. After applying LoRA, the number of parameters in DINOv2 was reduced from 1,100M to 5.91M, indicating a significant reduction in the computational cost of DINOv2-LoRA compared to DINOv2. Surprisingly, our LoRA fine-tuned model (DINOv2-LoRA) outperformed the fully fine-tuned original model (DINOv2) in most classes. This may be due to overfitting on our relatively small datasets when fine-tuning the entire large model like DINOv2 \citep{alzubaidi2021overfit}. In contrast, LoRA can potentially improve generalization by restricting the trainable parameters to low-rank adapters, thus mitigating the overfitting issues encountered by the fully fine-tuned original model \citep{ding2023overfit}. In summary, DINOv2-LoRA offers the best combination of accuracy and computational efficiency.

Apart from the Transformer-based models like DINOv2 and Swin-Transformer, EfficientNet-B7 achieved the best performance among those non-transformer-based models, achieving a match ratio of 58.34\%. Other models, including Swin-Transformer-Base, HRNet-W48, DenseNet-201, and ResNet-101, generated competitive performance in certain classes but were generally outperformed by Transformer-based models. Inception V3, proposed in 2016, achieved the worst performance among the tested models, reflecting the limitations of older architectures in tackling this complex ecological multi-label classification task. However, despite the less competitive results produced by conventional deep learning models compared to vision foundation models, this experiment establishes a benchmark for evaluating the superior performance of vision foundation models. These benchmarks serve as valuable reference points for future studies.

\begin{table}[ht]
    \centering
    \caption{Comparison of performance on the all-season dataset (\%) and the number of parameters (M) between baseline deep learning models and vision foundation models. The best performance is highlighted in bold, while the second highest is underlined. Abbreviations: HLC = Healthy Coral, CPC = Compromised Coral, DDC = Dead Coral, RBL = Rubble, CPT = Competition, DSE = Disease, PRD = Predation, and PHY = Physical Issues.}   
    \label{tab:mix_performance}
    \resizebox{1\linewidth}{!}{%
    \begin{tabular}{l l l l l l l l l l l l l}
        \hline
        \textbf{Model} & \textbf{Params.} & \textbf{Match Ratio} & \textbf{Micro F1} & \textbf{Macro F1} & \textbf{HLC} & \textbf{CPC} & \textbf{DDC} & \textbf{RBL} & \textbf{CPT} & \textbf{DSE} & \textbf{PRD} & \textbf{PHY}\\
        \hline
        VGG-19 & 143.7M & 51.33 & 81.34 & 73.72 & 93.26 & 72.36 & 81.20 & 66.71 & 71.89 & 72.43 & 73.09 & 58.82\\
        Inception V3 & 27.2M & 49.94 & 81.02 & 73.59 & 93.45 & 73.95 & 79.21 & 69.04 & 65.90 & 68.68 & 72.69 & 65.77\\
        ResNet-101 & 44.5M & 50.85 & 81.79 & 75.08 & 92.90 & 74.81 & 81.15 & 64.56 & 71.72 & 73.76 & 75.74 & 65.97\\
        DenseNet-201 & 20.0M & 54.72 & 83.64 & 77.87 & 94.16 & 76.95 & 82.28 & 72.82 & 74.95 & 76.89 & 77.92 & 66.94\\
        HRNet-W48 & 77.5M & 54.16 & 83.51 & 77.23 & 93.71 & 77.64 & 82.93 & 72.19 & 75.64 & 75.38 & 72.10 & 68.26\\
        EfficientNet-B7 & 66.3M & 58.34 & 85.23 & 80.34 & 94.91 & 79.75 & 83.44 & 73.34 & 77.40 & 78.69 & \underline{83.40} & 71.79\\
        Swin-Trans-Base & 87.8M & 60.34 & 85.86 & 80.42 & 95.18 & 80.11 & 85.27 & 74.75 & 78.95 & 78.53 & 80.88 & 69.72\\
        DINOv2 & 1100M & \underline{64.64} & \underline{87.83} & \underline{83.22} & \textbf{95.58} & \underline{82.63} & \underline{87.75} & \underline{78.83} & \textbf{83.69} & \underline{79.00} & 82.84 & \underline{75.42}\\
        DINOv2-LoRA & \textbf{5.91M} & \textbf{64.77} & \textbf{88.05} & \textbf{83.79} & \underline{95.57} & \textbf{83.01} & \textbf{87.77} & \textbf{79.27} & \underline{83.13} & \textbf{81.45} & \textbf{84.55} & \textbf{75.58}\\
        \hline
    \end{tabular}
    }
\end{table}

\begin{figure}[ht!]
\centering
\includegraphics[width=1\textwidth]{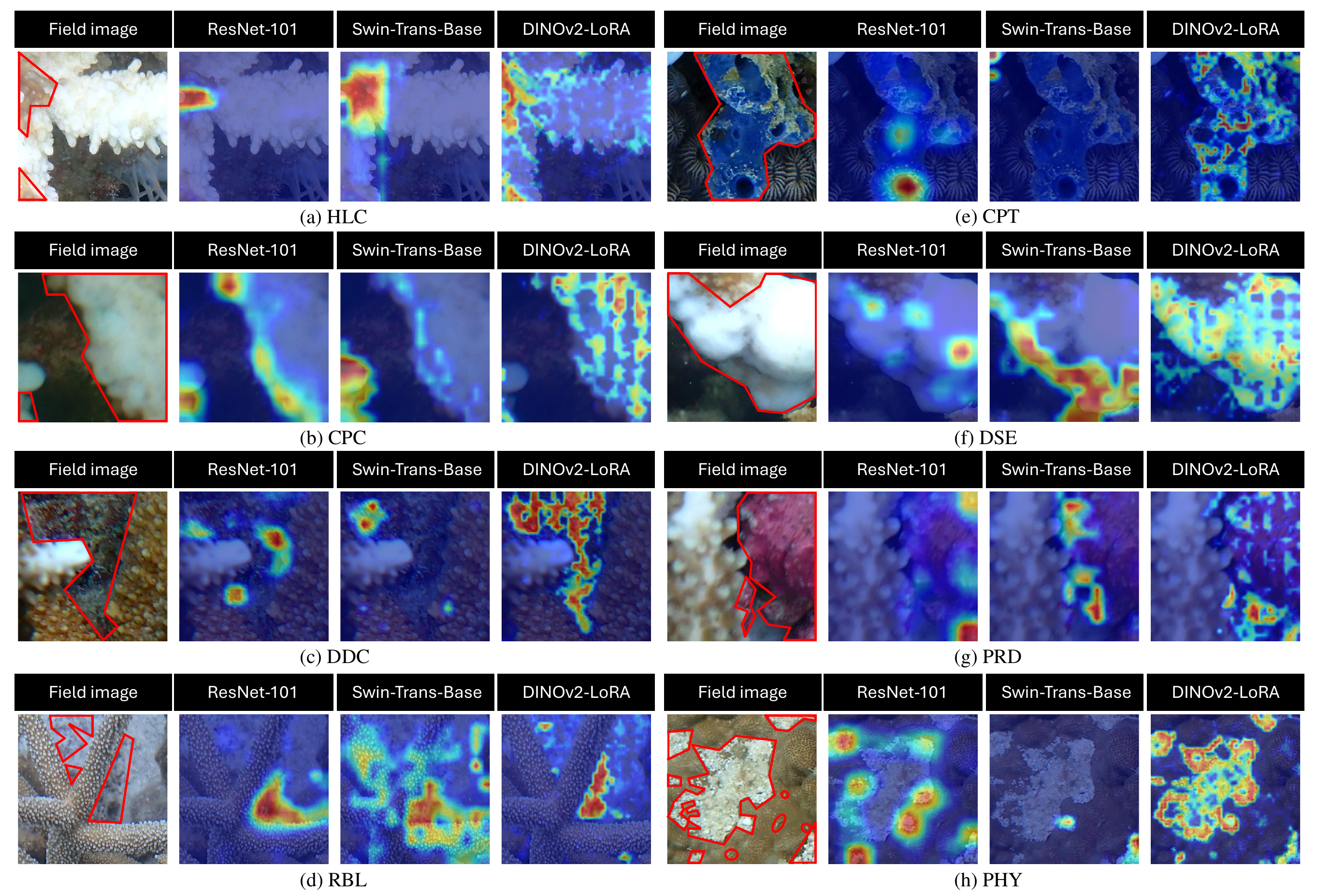} 
\caption{Grad-CAM heatmap visualizations for classification using ResNet-101, Swin-Transformer-Base, and DINOv2-LoRA on eight classes. The classes are: (a) Healthy Coral (HLC), (b) Compromised Coral (CPC), (c) Dead Coral (DDC), (d) Rubble (RBL), (e) Competition (CPT), (f) Disease (DSE), (g) Predation (PRD), and (h) Physical Issues (PHY). Targeted features representing the key areas the model should focus on are circled in red.}
\label{fig:grad-cam} 
\end{figure}

Figure \ref{fig:grad-cam} presents the Gradient-weighted Class Activation Mapping (Grad-CAM) \citep{Selvaraju2017Grad-CAM} heatmaps, providing a qualitative comparison of the performance of three representative models: ResNet-101, which is based on a convolutional neural network (CNN); Swin-Transformer-Base, which utilizes a transformer architecture; and DINOv2-LoRA, a vision foundation model. These Grad-CAM heatmaps visualize and highlight the regions most influential in the model's predictions, thereby identifying the model's attention. As shown in Figure \ref{fig:grad-cam}, only DINOv2-LoRA accurately highlights all areas containing the target features and identifies the precise boundaries of these features. In contrast, the heatmaps generated by the other two models either partially cover the target features or focus on background noise and irrelevant areas. For example, in Figure \ref{fig:grad-cam} (b) CPC, (c) DDC, (d) RBL, (e) CPT, (f) DSE, and (h) PHY, DINOv2-LoRA demonstrates an exact alignment between the coverage and boundaries of the target features in the field images and its Grad-CAM heatmaps. This illustrates the strong ability of DINOv2-LoRA to capture both the overall structure and subtle details of the target features. In contrast, for classes such as (d) RBL and (h) PHY, ResNet-101 captures only a portion of the target features, while Swin-Transformer-Base completely fails to locate the correct features. Overall, these Grad-CAM heatmaps demonstrate the precise and accurate attention of DINOv2-LoRA, validating its superior performance compared to the other two models. These findings of this qualitative comparison are consistent with the quantitative analyzes presented in Table \ref{tab:mix_performance}.

\subsection{Transfer learning capabilities of different models in multi-temporal set}
\label{sec:multi-temporal}
To evaluate the transfer learning capabilities of different models across seasons, their performance was tested under two experimental setups: (1) training the models on a dry season dataset (spring and summer) and transferring them to a wet season dataset (winter) and (2) training the models on a wet season dataset (winter) and transferring them to a dry season dataset (spring and summer). The results of these two experimental setups are presented in Table \ref{tab:seasonal_ss_to_w} and Table \ref{tab:seasonal_w_to_ss}, respectively.

According to Table \ref{tab:seasonal_ss_to_w}, significant performance differences can be observed between the validation set (dry season) and the test set (wet season). The performance drop across models provides insight into their ability to generalize across seasonal variations. For overall performance, DINOv2-LoRA achieved the best performance on the test set among all tested models. The model generated the highest Micro F1 score of 78.78\% and Macro F1 score of 67.07\%. It also showed superior performance for most classes, including Healthy Coral (HLC), Compromised Coral (CPC), Competition (CPT), Disease (DSE), and Physical Issues (PHY). Specifically, the model showed a significant improvement in the F1 scores in the Competition (CPT) and Physical Issues (PHY) classes, which are 4.61\% and 8.30\%, respectively. Although the F1 scores for the DINOv2-LoRA model are slightly lower than those of the DINOv2 model in Dead Coral (DDC), Rubble (RBL) and Predation (PRD), DINOV2-LoRA offers a more balanced overall performance when considering its computational efficiency. DINOv2-LoRA, with only 0.54\% of the trainable parameters of the original DINOv2 (5.91M compared to 1100M), achieved an optimal balance of transferability and accuracy across seasons, demonstrating its adaptability in this multi-label classification task under seasonal variations.

For performance retention across seasons, DINOv2-LoRA exhibited a smaller performance drop from validation to test set compared to other models, with the Match Ratio decreasing from 66.11\% to 46.63\%. This indicates relatively strong transferability across seasons. Similarly, DINOv2 demonstrated robust transferability; however, its performance drop was slightly greater than that of DINOv2-LoRA in terms of F1 scores and class-level metrics. Conventional deep learning models such as DenseNet-201, ResNet-101, and HRNet-W48 performed well on the validation set. For example, EfficientNet-B7 (V) yielded a match ratio of 60.02\%. However, when evaluated on the test set, these models experienced substantial drops across all metrics, indicating low seasonal transferability. These significant performance drops highlight the limitations of conventional deep learning models in handling this complex ecological multi-label classification task. In summary, these findings demonstrate that DINOv2-LoRA offers the highest transferability and efficiency among all tested models, making it suitable for tasks involving seasonal variability.

\begin{table}[ht]
    \centering
    \caption{Performance comparison of transfer learning capabilities across seasons, from the dry season (validation set) to the wet season (test set), using various models. The best performance on the test set is in bold, while the second highest is underlined. Abbreviations: V = model's performance on the validation set, T = model's performance on the test set, HLC = Healthy Coral, CPC = Compromised Coral, DDC = Dead Coral, RBL= Rubble, CPT = Competition, DSE = Disease, PRD = Predation, and PHY = Physical Issues.}   
    \label{tab:seasonal_ss_to_w}
    \resizebox{1\linewidth}{!}{%
    \begin{tabular}{l l l l l l l l l l l l}
        \hline
        \textbf{Model} & \textbf{Match Ratio} & \textbf{Micro F1} & \textbf{Macro F1} & \textbf{HLC} & \textbf{CPC} & \textbf{DDC} & \textbf{RBL} & \textbf{CPT} & \textbf{DSE} & \textbf{PRD} & \textbf{PHY}\\
        \hline
        VGG-19 (V) & 51.48 & 81.32 & 74.81 & 93.34 & 73.14 & 79.28 & 64.96 & 73.47 & 72.34 & 79.10 & 62.85\\
        Inception V3 (V) & 49.98 & 80.84 & 74.46 & 93.37 & 73.04 & 77.93 & 68.24 & 68.21 & 72.25 & 77.55 & 65.08\\
        ResNet-101 (V) & 53.18 & 82.66 & 76.64 & 93.74 & 76.29 & 81.21 & 62.99 & 74.51 & 75.85 & 84.07 & 64.48\\
        DenseNet-201 (V) & 54.86 & 83.63 & 77.79 & 94.34 & 78.55 & 81.24 & 68.01 & 74.91 & 75.77 & 82.44 & 67.06\\
        HRNet-W4 8(V) & 56.37 & 83.94 & 78.36 & 94.20 & 78.65 & 82.05 & 69.89 & 78.15 & 76.38 & 86.33 & 61.24\\
        EfficientNet-B7 (V) & 60.02 & 85.54 & 80.53 & 95.29 & 80.12 & 83.51 & 71.45 & 79.04 & 79.14 & 85.64 & 70.01\\
        Swin-Trans-Base (V) & 62.24 & 86.52 & 81.67 & 95.10 & 81.82 & 85.25 & 74.10 & 81.62 & 80.42 & 84.18 & 70.83\\
        DINOv2 (V) & 65.98 & 88.28 & 83.90 & 95.84 & 84.30 & 87.25 & 77.40 & 83.21 & 83.59 & 86.03 & 73.60\\
        DINOv2-LoRA (V) & 66.11 & 88.14 & 83.79 & 95.84 & 84.35 & 87.28 & 75.10 & 83.02 & 82.37 & 89.07 & 73.31\\
        VGG-19 (T) & 32.61 & 68.43 & 52.22 & 90.05 & 48.87 & 65.14 & 55.51 & 36.09 & 39.51 & 0.36 & 46.62\\
        Inception V3 (T) & 32.85 & 69.83 & 53.67 & 90.30 & 52.84 & 67.21 & 58.33 & 30.07 & 37.97 & 40.96 & 51.70\\
        ResNet-101 (T) & 31.28 & 69.47 & 53.40 & 89.94 & 56.70 & 68.28 & 39.60 & 28.84 & 48.22 & 47.67 & 48.25\\
        DenseNet-201 (T) & 33.30 & 70.64 & 55.15 & 91.62 & 57.09 & 68.30 & 45.57 & 31.61 & 42.76 & 51.87 & 52.37\\
        HRNet-W4 8(T) & 34.77 & 70.60 & 53.24 & 91.20 & 53.80 & 69.91 & 51.87 & 31.43 & 42.76 & 44.67 & 40.28\\
        EfficientNet-B7 (T) & 37.35 & 72.31 & 57.82 & 91.73 & 58.90 & 69.08 & 55.02 & 34.02 & 49.45 & 55.38 & 48.98\\
        Swin-Trans-Base (T) & 36.97 & 72.01 & 55.99 & 90.20 & 58.20 & 72.74 & 52.20 & 38.55 & 42.67 & 43.31 & 50.00\\
        DINOv2 (T) & \textbf{46.90} & \underline{78.60} & \underline{65.47} & \underline{93.78} & \underline{63.25} & \textbf{81.50} & \textbf{64.51} & \underline{44.75} & \underline{54.22} & \textbf{64.26} & \underline{57.51}\\
        DINOv2-LoRA (T) & \underline{46.63} & \textbf{78.78} & \textbf{67.07} & \textbf{93.89} & \textbf{65.75} & \underline{78.46} & \underline{63.91} & \textbf{49.36} & \textbf{56.48} & \underline{62.91} & \textbf{65.81}\\
        \hline
    \end{tabular}
    }
\end{table}

The results presented in Table \ref{tab:seasonal_w_to_ss} reflect the performance of various models when trained on a wet season dataset (validation set) and tested on a dry season dataset (test set). This experimental setup complements the first setup by evaluating transfer learning capabilities in the reverse seasonal direction. For overall performance, DINOv2-LoRA again achieved the highest accuracy, with a micro F1 score (73.84\%) and a macro F1 score (60.52\%) on the test set. It outperformed all other models in most classes, including Healthy Coral (HLC), Dead Coral (DDC), and Competition (CPT), demonstrating its adaptability to seasonal changes. For performance retention across seasons, the performance drops in the wet-to-dry transfer setting were more pronounced than in the dry-to-wet transfer setting for all models. This suggests that seasonal variability has a greater impact when moving from wet to dry seasons. Specifically, the performance of conventional deep learning models such as EfficientNet-B7, DenseNet-201, and HRNet-W48 on the validation set experienced significant performance drops when transferred to the test set. For example, the match ratio of EfficientNet-B7 dropped from 64.73\% on the validation set to 27.88\% on the test set, reflecting challenges in generalizing across seasons. In summary, the results derived from this experimental setup are consistent with the findings of the first setup, with vision foundation models (DINOv2-LoRA and DINOv2) outperforming conventional deep learning architectures in accuracy and transferability.

\begin{table}[ht]
    \centering
    \caption{Performance comparison of transfer learning capabilities across seasons, from the wet season (validation set) to the dry season (test set), using various models. The best performance is in bold, while the second highest is underlined. Abbreviations: V = model's performance on the validation set, T = model's performance on the test set, HLC = Healthy Coral, CPC = Compromised Coral, DDC = Dead Coral, RBL= Rubble, CPT = Competition, DSE = Disease, PRD = Predation, and PHY = Physical Issues.}   
    \label{tab:seasonal_w_to_ss}
    \resizebox{1\linewidth}{!}{%
    \begin{tabular}{l l l l l l l l l l l l}
        \hline
        \textbf{Model} & \textbf{Match Ratio} & \textbf{Micro F1} & \textbf{Macro F1} & \textbf{HLC} & \textbf{CPC} & \textbf{DDC} & \textbf{RBL} & \textbf{CPT} & \textbf{DSE} & \textbf{PRD} & \textbf{PHY}\\
        \hline
        VGG-19 (V) & 57.81 & 85.32 & 77.43 & 95.40 & 77.91 & 85.18 & 78.83 & 77.69 & 71.46 & 59.65 & 73.36\\
        ResNet-101 (V) & 58.92 & 86.77 & 80.03 & 95.89 & 79.62 & 86.18 & 79.49 & 80.08 & 76.82 & 64.79 & 77.37\\
        Inception V3 (V) & 55.76 & 84.85 & 76.62 & 94.94 & 77.46 & 84.57 & 78.10 & 75.98 & 74.47 & 53.73 & 73.70\\
        DenseNet-201 (V) & 61.36 & 87.57 & 81.90 & 96.25 & 81.21 & 86.73 & 81.88 & 82.80 & 77.09 & 72.13 & 77.12\\
        HRNet-W48 (V) & 61.19 & 87.48 & 82.07 & 95.77 & 82.63 & 86.56 & 80.66 & 82.10 & 79.58 & 74.19 & 75.09\\
        EfficientNet-B7 (V) & 64.73 & 88.43 & 82.92 & 96.22 & 81.95 & 88.64 & 83.30 & 84.93 & 79.66 & 71.64 & 77.05\\
        Swin-Trans-Base (V) & 66.82 & 89.68 & 84.59 & 96.29 & 84.76 & 89.53 & 84.40 & 87.33 & 83.87 & 71.23 & 79.32\\
        DINOv2 (V) & 70.20 & 90.72 & 85.90 & 97.26 & 87.27 & 90.03 & 82.94 & 87.55 & 83.52 & 77.97 & 80.63\\
        DINOV2-LoRA (V) & 71.14 & 91.26 & 86.49 & 97.12 & 87.26 & 90.32 & 88.06 & 90.15 & 85.09 & 72.41 & 81.53\\
        VGG-19 (T) & 22.76 & 63.37 & 47.19 & 82.48 & 56.96 & 66.95 & 36.79 & 26.95 & 41.57 & 29.76 & 36.03\\
        ResNet-101 (T) & 21.17 & 64.19 & 45.65 & 84.53 & 55.40 & 66.66 & 25.90 & 22.50 & 44.90 & 38.43 & 26.90\\
        Inception V3 (T) & 22.05 & 63.99 & 46.92 & 84.14 & 56.88 & 66.93 & 36.59 & 29.96 & 42.59 & 24.88 & 33.41\\
        DenseNet-201 (T) & 23.61 & 64.54 & 48.63 & 84.81 & 56.34 & 66.17 & 34.47 & 32.90 & 37.36 & 43.63 & 33.33\\
        HRNet-W48 (T) & 23.33 & 64.51 & 50.07 & 84.81 & 56.89 & 66.15 & 39.17 & 37.76 & 43.03 & 41.68 & 31.07\\
        EfficientNet-B7 (T) & 27.88 & 68.02 & 54.20 & 86.31 & 60.99 & 70.10 & 44.17 & 39.22 & 47.86 & 42.15 & 42.84\\
        Swin-Trans-Base (T) & 25.29 & 66.64 & 51.95 & 85.68 & 58.55 & 69.89 & 31.86 & 36.58 & 47.15 & \underline{49.20} & 36.68\\
        DINOv2 (T) & \textbf{35.42} & \underline{72.66} & \underline{58.64} & \underline{88.77} & \textbf{66.34} & \underline{74.32} & \underline{46.19} & \underline{41.79} & \underline{52.64} & 48.18 & \underline{50.90}\\
        DINOV2-LoRA (T) & \underline{35.38} & \textbf{73.84} & \textbf{60.52} & \textbf{90.37} & \underline{66.16} & \textbf{75.04} & \textbf{47.52} & \textbf{43.07} & \textbf{58.20} & \textbf{49.12} & \textbf{54.68}\\
        \hline
    \end{tabular}
    }
\end{table}

\subsection{Transfer learning capabilities of different models in multi-spatial set}
\label{sec:multi-spatial}
Apart from evaluating multi-temporal transferability, this study also assessed the transferability of different models across various habitats. Specifically, images collected from TTB, ALK, SKI, and CBK were used as the test set, while images from the remaining 11 habitats were used for training and validation. The results are presented in Table \ref{tab:multispatial_performance}. For overall performance, DINOv2-LoRA consistently outperformed other models on the test set, achieving the highest Match Ratio, Micro F1 score, and Macro F1 score. Although DINOv2-LoRA performed slightly below DINOv2 in specific classes such as Dead Coral (DDC) (1.34\% lower) and Competition (CPT) (0.84\% lower), it provided a better balance between performance and computational efficiency, making it the optimal choice for multi-spatial transfer learning tasks.

For performance retention across habitats, all tested models exhibited significant performance drops from validation to test set. Among these, DINOv2-LoRA showed the smallest drop, with the match ratio decreasing by less than 20\%, while all conventional deep learning models experienced declines greater than 20\%. This result highlights the relatively robust multi-spatial transferability of DINOv2-LoRA among the tested models. Similarly, DINOv2 showed the second-best transferability, with a 20.84\% decline in the match ratio. However, its overall performance drop was slightly greater than that of DINOv2-LoRA in most evaluation metrics. Conventional deep learning models achieved moderate performance on the validation set, but suffered significant drops on the test set. For instance, DenseNet-201 achieved a match ratio of 55.21\% on the validation set, which dropped to 33.69\% on the test set, reflecting its limited ability to adapt to unseen habitats. Similarly, the best conventional model, EfficientNet-B7, had a match ratio drop from 59.86\% to 39.07\%, indicating moderate transferability, but still lagged behind transformer-based models.

In conclusion, the findings in this section highlight that DINOv2-LoRA offers exceptional transfer learning capabilities across different habitats, outperforming other models in accuracy and efficiency. The consistent superiority of transformer-based architectures in this study reinforces their effectiveness in addressing spatial variability in ecological multi-label classification tasks.

\begin{table}[ht]
    \centering
    \caption{Performance comparison of transfer learning across different habitats using various models. The best performance is in bold, while the second highest is underlined. Abbreviations: HLC = Healthy Coral, CPC = Compromised Coral, DDC = Dead Coral, RBL= Rubble, CPT = Competition, DSE = Disease, PRD = Predation, and PHY = Physical Issues.}   
    \label{tab:multispatial_performance}
    \resizebox{1\linewidth}{!}{%
    \begin{tabular}{l l l l l l l l l l l l}
        \hline
        \textbf{Model} & \textbf{Match Ratio} & \textbf{Micro F1} & \textbf{Macro F1} & \textbf{HLC} & \textbf{CPC} & \textbf{DDC} & \textbf{RBL} & \textbf{CPT} & \textbf{DSE} & \textbf{PRD} & \textbf{PHY}\\
        \hline
        VGG-19 (V) & 51.43 & 81.29 & 74.78 & 93.28 & 72.44 & 80.33 & 65.07 & 69.44 & 72.07 & 80.39 & 65.23\\
        Inception V3 (V) & 51.36 & 81.53 & 74.73 & 93.54 & 73.55 & 80.51 & 66.72 & 66.33 & 69.96 & 79.40 & 67.82\\
        ResNet-101 (V) & 53.70 & 82.82 & 76.55 & 93.85 & 75.43 & 82.84 & 66.37 & 70.14 & 73.60 & 81.45 & 68.75\\
        DenseNet-201 (V) & 55.21 & 84.00 & 77.82 & 94.34 & 76.83 & 83.85 & 69.57 & 73.94 & 74.73 & 79.12 & 70.21\\
        HRNet-W48 (V) & 55.13 & 83.63 & 77.84 & 93.49 & 77.76 & 83.95 & 68.64 & 71.48 & 75.40 & 82.23 & 69.73\\
        EfficientNet-B7 (V) & 59.86 & 85.59 & 80.39 & 94.89 & 79.75 & 85.39 & 71.91 & 77.36 & 78.12 & 84.91 & 70.80\\        
        Swin-Trans-Base (V) & 62.03 & 86.63 & 81.34 & 95.52 & 79.54 & 87.06 & 72.01 & 78.34 & 81.21 & 83.73 & 73.27\\
        DINOv2 (V) & 66.01 & 88.40 & 84.07 & 95.74 & 83.82 & 88.31 & 75.90 & 82.96 & 82.23 & 87.47 & 76.08\\
        DINOV2-LoRA (V) & 66.07 & 88.58 & 84.82 & 96.04 & 83.58 & 88.21 & 77.00 & 81.81 & 82.66 & 90.78 & 78.48\\        
        VGG-19 (T) & 33.58 & 69.32 & 56.51 & 91.12 & 58.84 & 56.32 & 45.01 & 57.05 & 55.53 & 49.12 & 39.10\\
        Inception V3 (T) & 33.56 & 69.00 & 55.91 & 90.94 & 60.28 & 59.56 & 45.75 & 47.10 & 54.59 & 48.63 & 40.48\\
        ResNet-101 (T) & 31.90 & 68.09 & 56.12 & 87.90 & 60.98 & 63.35 & 43.95 & 51.29 & 49.09 & 50.91 & 41.46\\
        DenseNet-201 (T) & 33.69 & 70.72 & 58.59 & 90.21 & 62.10 & 63.70 & 46.19 & 59.58 & 52.52 & 50.00 & 44.41\\
        HRNet-W48 (T) & 31.77 & 68.72 & 56.32 & 90.36 & 60.66 & 61.34 & 43.84 & 48.21 & 54.18 & 46.40 & 45.58\\
        EfficientNet-B7 (T) & 39.07 & 73.24 & 62.76 & 91.23 & 66.06 & 66.49 & 53.85 & 60.28 & 58.77 & 60.81 & 44.59\\
        Swin-Trans-Base (T) & 38.98 & 73.42 & 63.06 & 91.55 & 65.29 & 65.79 & 43.87 & \underline{65.09} & 58.11 & 63.06 & 51.71\\
        DINOv2 (T) & \underline{45.17} & \underline{77.25} & \underline{67.29} & \underline{93.74} & \underline{72.09} & \textbf{70.11} & \underline{51.75} & \textbf{65.49} & \textbf{64.15} & \underline{65.00} & \underline{55.97}\\
        DINOV2-LoRA (T) & \textbf{46.92} & \textbf{77.92} & \textbf{69.42} & \textbf{94.37} & \textbf{73.18} & \underline{68.77} & \textbf{59.07} & 64.65 & \underline{62.13} & \textbf{68.75} & \textbf{64.46}\\
        \hline
    \end{tabular}
    }
\end{table}

\subsection{Effects of the low-rank number on the model's performance}
\begin{figure}[ht]
\centering
\includegraphics[width=0.9\textwidth]{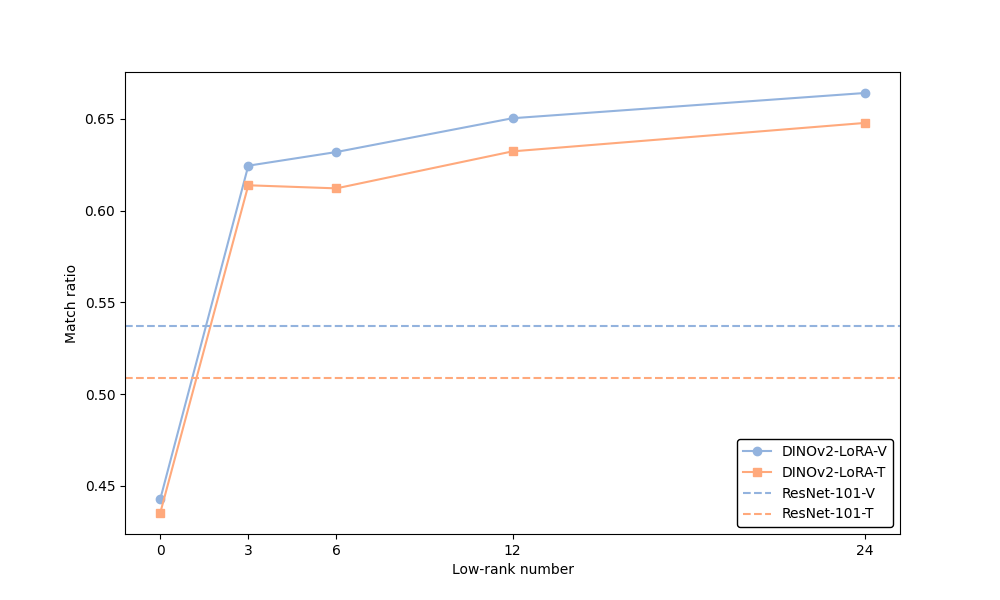} 
\caption{The effects of the low-rank number ($r$) on the match ratio. ``V" refers to the validation set, and ``T" refers to the test set. ResNet-101 (44.5M parameters) is used as the baseline to evaluate the computational efficiency of the proposed DINOv2-LoRA model (5.91M parameters when $r$ is 24).}
\label{fig:low_rank_effect} 
\end{figure}

Given the trade-off between model performance and computational efficiency, this study evaluated a range of low-rank values ($r$) to determine the optimal setting. As shown in Figure \ref{fig:low_rank_effect}, $r=0$ corresponds to a frozen DINOv2 backbone, where only the classifier is fine-tuned. In this configuration, the match ratio is the lowest, indicating that the pre-trained knowledge of DINOv2 alone is insufficient for tasks requiring highly specialized domain expertise, such as coral reef condition classification. This result underscores the importance of leveraging LoRA to fine-tune both the classifier and the LoRA layers, enabling DINOv2 to adapt effectively to this domain-specific task. As $r$ increases, the match ratio for both DINOv2-LoRA-V and DINOv2-LoRA-T improves substantially. However, performance gains diminish as $r$ approaches 24, indicating a plateau in improvement. This trend suggests that $r=24$ represents the optimal value to balance model performance with computational efficiency. Meanwhile, the match ratio generated by ResNet-101 on the validation and test sets is plotted in Figure \ref{fig:low_rank_effect} as a baseline to compare the computational costs saved by the LoRA adapter. Specifically, ResNet-101, with 44.5M parameters, achieved a match ratio of 53.49\% on the validation set, while DINOv2-LoRA achieved a match ratio of 62.44\% with $r=3$. This finding suggests that DINOv2-LoRA can achieve promising performance by fine-tuning only a small number of trainable parameters (i.e., with a very small rank $r$ of 3), indicating the significant computational costs saved by adapting LoRA for fine-tuning DINOv2.

\section{Discussion}
\subsection{Analyzing the misclassified images}
\begin{figure}[ht!]
\centering
\includegraphics[width=0.9\textwidth]{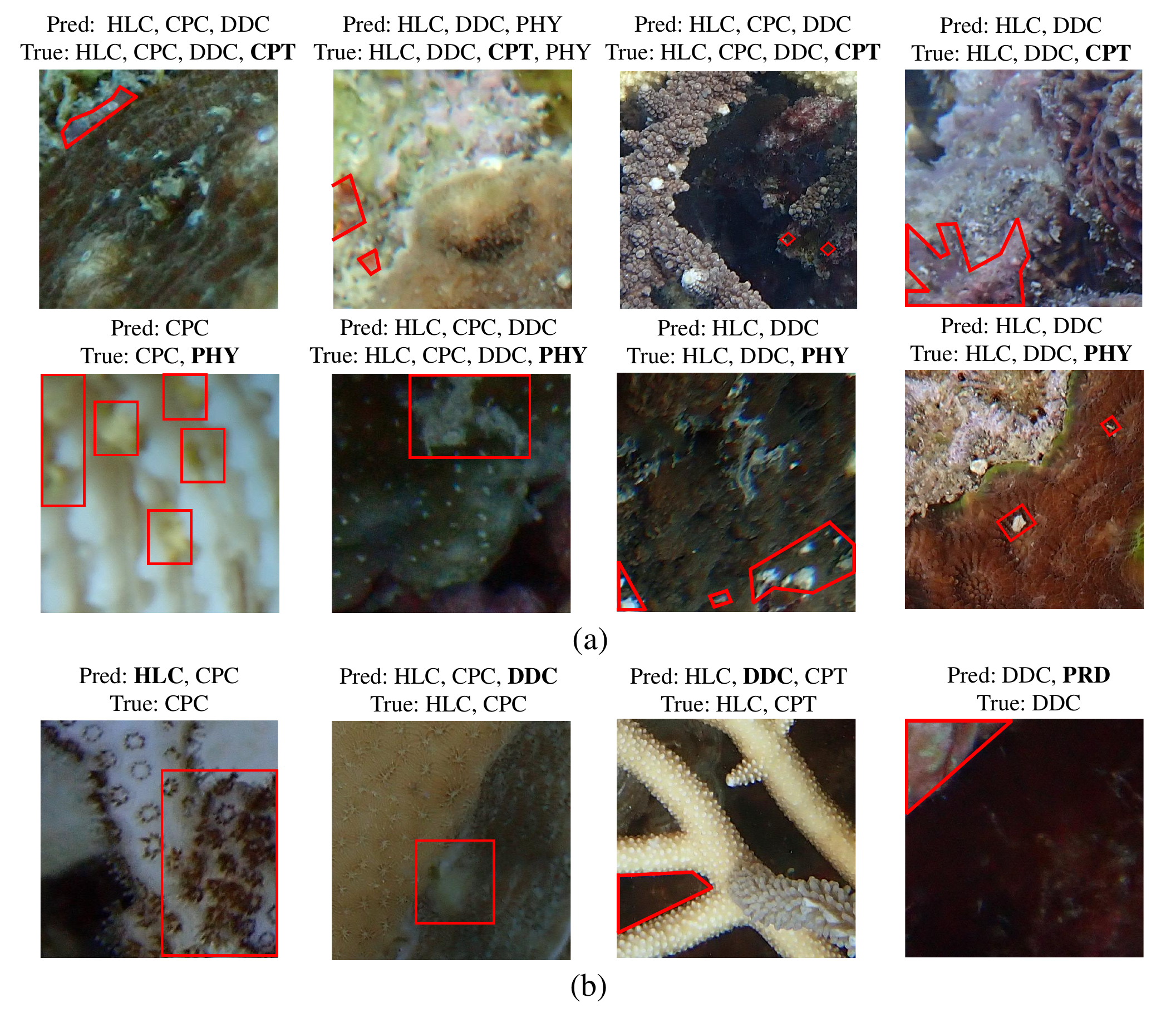} 
\caption{Examples of misclassified images by DINOv2-LoRA on the all-season dataset. Incorrect labels are highlighted in bold, and missed features that the model should have focused on are circled in red. (a) depicts false negatives, where the model failed to detect the presence of present instances; (b) depicts false positives, where the model incorrectly identified instances as being present when they are not.}
\label{fig:misclassified_samples} 
\end{figure}

Apart from the quantitative analysis of the performance of the proposed DINOv2-LoRA model presented in Table \ref{tab:mix_performance}, this section examines misclassified images to provide additional qualitative insight into the performance of the model and the challenges it faces. Figure \ref{fig:misclassified_samples} illustrates 12 misclassified samples, along with their incorrect predicted labels and the corresponding ground truth labels. Figure \ref{fig:misclassified_samples} (a) illustrates false negatives, where the model failed to detect instances that are actually present. For example, images in row 1, columns 1–3, depict small target areas indicating the presence of class Competition (CPT) such as cyanobacteria and macroalgae. Meanwhile, the image in row 1, column 4, shows cyanobacteria with a color similar to adjacent corals, which the model failed to identify. The images in row 2 further demonstrate the model's failure to identify subtle instances belonging to the Physical Issues class. These findings suggest that the model's sensitivity to subtle features of the Competition (CPT) and Physical Issues (PHY) classes is inadequate, likely due to the relatively small sample size of these two classes \citep{Leevy2018imbalance}. Figure \ref{fig:misclassified_samples} (b) illustrates false positives, where the model incorrectly identified instances as being present when they are not. For example, the image in row 3, column 1 contains brown coral polyps that are characteristic of the class Healthy Coral (HLC). However, most tissues are missing, and the white skeleton is clearly visible in the image, indicating that it should be labeled as Compromised Coral (CPC) only. Similarly, features similar to those of the Dead Coral (DDC) or Predation (PRD) classes are observed in images in row 3, columns 2–4, which confuse the model and lead to incorrect predictions of the presence of these two classes. Further work is recommended to address these issues, including applying techniques to mitigate class imbalance \citep{wu2020imbalance}, incorporating more representative samples from minority classes \citep{Wong2016minority}, and integrating modules that are sensitive to global structural information rather than relying solely on local texture and color information \citep{He2022globalfeature}.

\subsection{Variations across different seasons and habitats}
Since significant performance drops were observed when transferring the models across different seasons and habitats, this section discusses variations in the images' characteristics across different seasons and habitats to account for the performance drops.

\begin{figure}[ht!]
    \centering
    % First subfigure
    \begin{subfigure}[b]{0.48\linewidth}
        \centering
        \includegraphics[width=\linewidth]{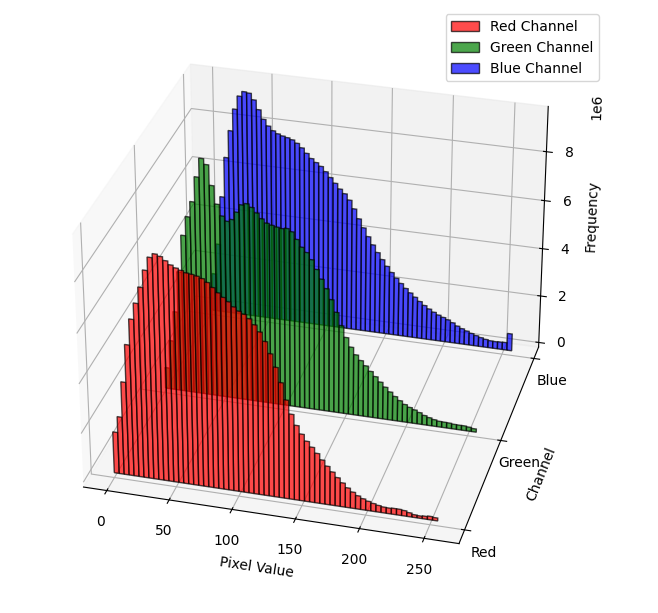}
        \caption{}
        \label{fig:dry_histogram}
    \end{subfigure}
    \hfill
    % Second subfigure
    \begin{subfigure}[b]{0.48\linewidth}
        \centering
        \includegraphics[width=\linewidth]{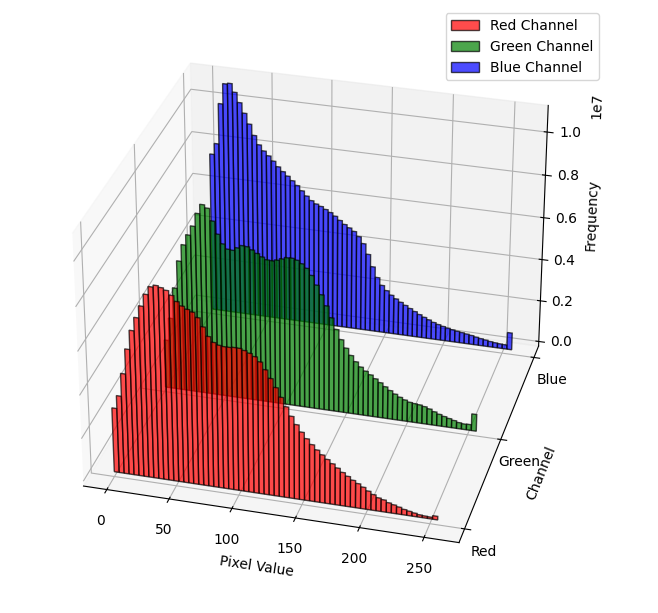}
        \caption{}
        \label{fig:wet_histogram}
    \end{subfigure}
    \caption{Pixel value histograms of image samples (N=1038) for (a) the dry season dataset and (b) the wet season dataset.}
    \label{fig:seasonal_diff_histogram}
\end{figure}

The seasonal variations are due to the change in image quality and ecological traits. For image quality, the underwater environment becomes less favorable for taking photos during the wet season due to the decrease in light intensity and water clarity, increased turbidity, greater differences between surface and underwater temperature, stronger water currents, and blooms of marine phytoplankton \citep{Bordalo2001waterquality}. Specifically, increased turbidity and decreased water clarity can lead to blurred images. The increase in nutrients in seawater brought by the land runoff from heavy precipitation can cause algae blooms, resulting in noise in collected images \citep{Diego-Feliu2022rainfall}. The reduced light intensity makes it more challenging to achieve proper white balance without additional artificial lighting, leading to color distortion that gives the images a bluish or greenish tint \citep{Ancuti2018colour}. Figure \ref{fig:seasonal_diff_histogram} illustrates how color distortion can be quantified. A total of 1038 image patches were randomly selected from both the dry and wet seasons, and their pixel value histograms were calculated for the red, green, and blue channels. The frequency of pixels in the blue channel is relatively higher than in the green and red channels, indicating the dominant "blue" appearance in the images. Consequently, the image quality of the wet season dataset is lower than that of the dry season dataset. This explains why, as shown in Table \ref{tab:seasonal_w_to_ss} and Table \ref{tab:seasonal_ss_to_w}, the overall accuracy of the models trained on the wet season dataset is lower than that of the models trained on the dry season dataset. Furthermore, there is a more significant performance drop when transferring models from the wet-season dataset to the dry-season dataset compared to the reverse experimental setting. 

\begin{figure}[ht!]
\centering
\includegraphics[width=0.9\textwidth]{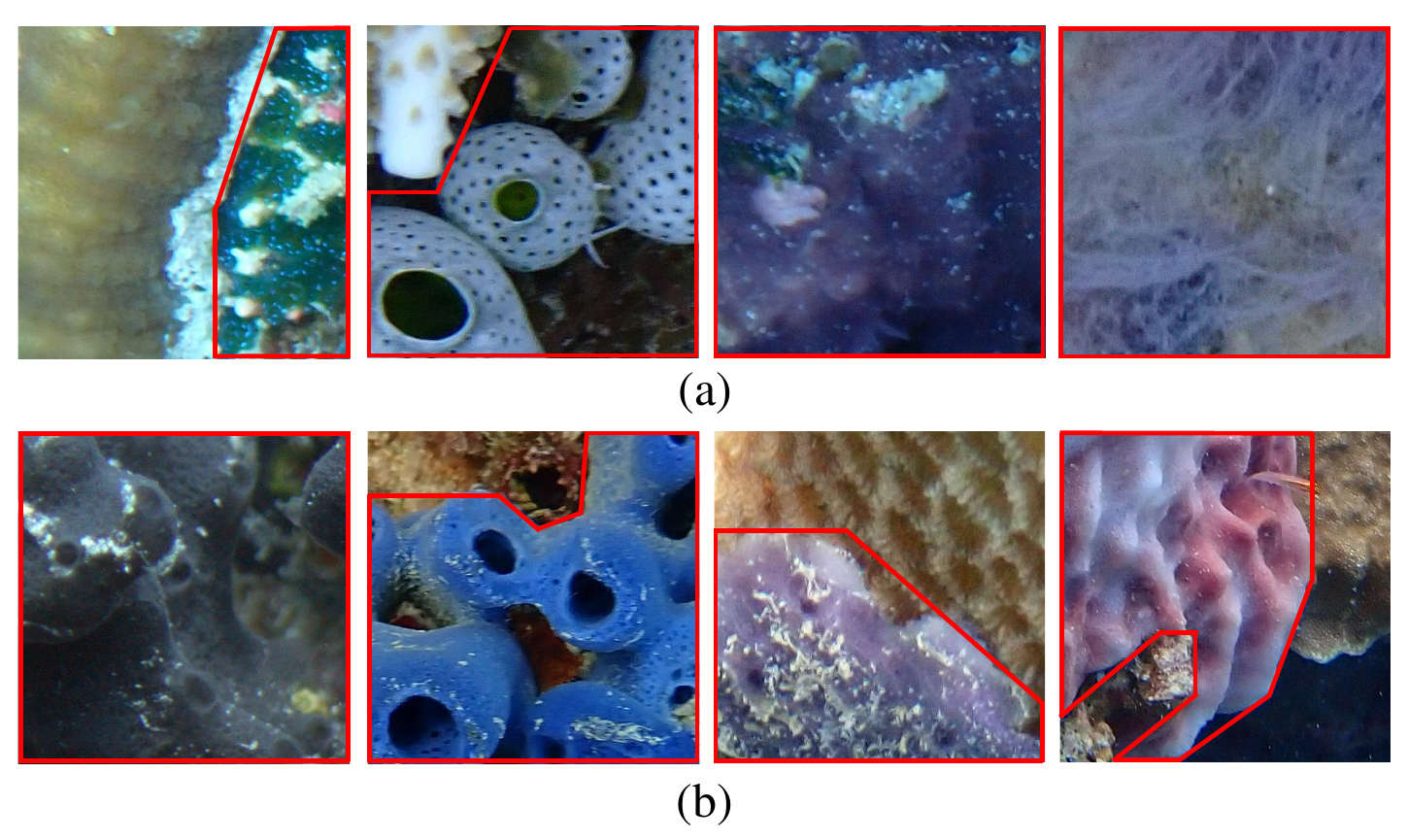} 
\caption{(a) Examples of dominant competitors, i.e., tunicate and cyanobacteria, in the dry season dataset. (b) Examples of dominant competitors, i.e., sponge, from class competition (CPT) in the wet season dataset.}
\label{fig:seasonal_variations} 
\end{figure}

Changes in the ecological traits of the instances captured in the images contribute to the seasonal variations between the dry and wet datasets. Coral reef ecosystems are highly dynamic, and coral conditions and flora change in response to seasonal variations \citep{harrison2007seasondynamic}. For example, the water temperature is lower during the wet season compared to the dry season, resulting in a reduced heat stress on corals. Consequently, the rates of disease and coral bleaching are generally lower during the wet season than during the dry season \citep{Brodnicke2019heat}, leading to different class distribution patterns between the two datasets. This distribution difference between the datasets contributes to the performance drop observed in cross-season transferability experiments \citep{day2017distribution}. In addition to different data distributions, changes in dominant competitors also lead to differences in traits between dry and wet season datasets. As shown in Figure \ref{fig:seasonal_variations}, cyanobacteria and tunicates are dominant competitors during the calm and warm dry season \citep{Kanoshina2003cyanobacterial}, while both decrease or even disappear and sponges become dominant competitors during the wet season \citep{aerts1997sponge}. Consequently, the traits of the images differ between the wet and dry season datasets, making it challenging for a model trained on one dataset to transfer effectively to the other.

Habitat variations are influenced by changes in the physical and biological environments of different diving sites. Specifically, the depth of each sampled habitat ranges from 0.3 to 23.3 m (detailed geographical information is provided in the Data availability statement). The topography also varies between habitats, including reef crests, patch reefs, underwater pinnacles, and reef flats. Therefore, the composition of coral species and the growth forms of coral colonies differ between habitats \citep{done1999spatialvariation}, resulting in different traits in the images collected in different habitats. These variations in traits across habitats contribute to the performance drop observed in the spatial transferability experiments.

\subsection{Implications for coral reef conservation}
The DINOv2-LoRA model proposed in this study offers an automatic, accurate, and efficient tool for qualitative analysis of coral reef conditions. This model enables the classification of coral reef images into health statuses (healthy, compromised, dead, and rubble) and the corresponding stressors (competition, disease, predation, and physical issues). In line with the universal classification standard used in citizen science-based conservation programs \citep{scott2019ecological}, the proposed DINOv2-LoRA model significantly improves efficiency by reducing the time required to label an image from five seconds (when done by a trained volunteer) to 0.02 seconds (when processed by the trained model). In addition, inexperienced volunteers are prone to errors, whereas DINOv2-LoRA achieved a micro F1 score of 88.05\% and a macro F1 score of 83.79\% on the all-season dataset. Although few studies have addressed similar tasks, these F1 scores are relatively high (greater than 80\%) compared to other multi-label classification tasks based on ecological data \citep{Swaminathan2024f1score}, demonstrating the robust performance of the proposed model.

\begin{figure}[ht!]
\centering
\includegraphics[width=0.9\textwidth]{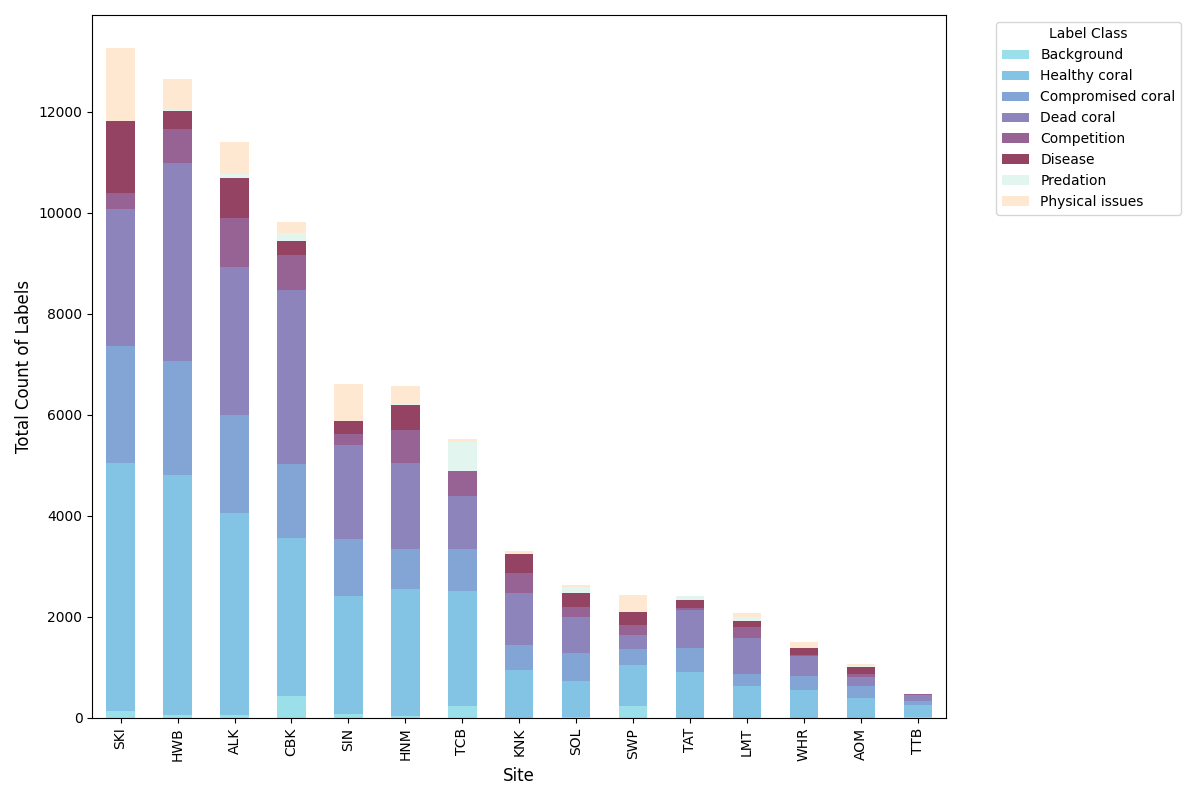} 
\caption{Label frequency and composition at each site. }
\label{fig:site_specific_label} 
\end{figure}

After the coral reef images are classified by the proposed model, ecological information can be extracted to guide conservation activities. As shown in Figure \ref{fig:site_specific_label}, the frequency and composition of the labels at each site reflect the status of the habitat. For instance, the proportion of the class Predation is relatively high compared to other classes, suggesting a potential population outbreak of predators at this site \citep{zhang2024drupella}. Consequently, conservation activities such as predator removal are recommended to mitigate predator damage and protect coral colonies in the area. Similarly, the high proportion of the class Physical Issues at SIN and SWP indicates the need for solutions like installing sediment traps and reforestation in coastal regions \citep{Storlazzi2011sediment}. However, it should be noted that this study aims to collect representative images to train the classification model to aid the qualitative analysis of coral reef conditions, a process consistent with the current approach used by our local NPO team. Consequently, the number of images collected at each site varies, and the size of each image is not uniform, which means that the proposed model is not suitable for quantitative analysis. Future work could focus on using images collected with quadrat and applying semantic segmentation models if quantitative analysis is required.

\subsection{Limitations and future work}
\begin{figure}[ht!]
\centering
\includegraphics[width=0.7\textwidth]{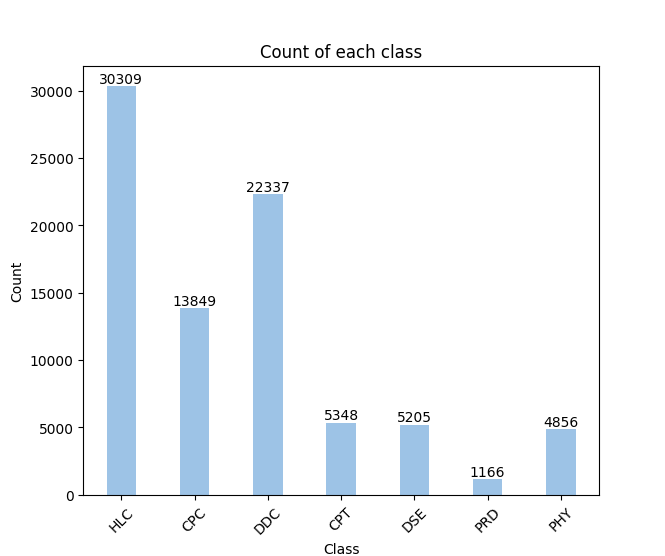} 
\caption{Imbalanced distribution of the dataset.}
\label{fig:class_distribution} 
\end{figure}

The dataset used in this study is imbalanced, as shown in Figure \ref{fig:class_distribution}, where the the labels are distributed non-uniformly across classes, with the class Healthy Coral (HLC) comprising the highest number of labels. This is due to the inherent characteristics of ecological data; in stable coral reef habitats, healthy corals should dominate, while compromised corals and stressors are less common \citep{sorokin2013coralecology}. Otherwise, habitats are at risk of collapse. However, an imbalanced dataset can lead to a performance drop in less common classes, and an effective model should accurately label these minority classes \citep{Tarekegn2021imbalance}. Some studies have applied data augmentation methods to address the issue of class imbalance. For example, \citet{Ibrahim2018augmentation, rana2022augmentation, escobar2024augmentation} proposed deep learning-based data augmentation techniques to generate additional samples for minority classes, thus enriching the features learned by the model and improving the performance of classification models on these rare classes \citep{Jiang2021augmentation}. However, as the end users of our model are staff members from conservation communities or government sectors with limited expertise in computer vision, extra pre-processing steps like data augmentation can complicate this classification task. Therefore, to make the proposed process and model as simple and user-friendly as possible, we chose not to apply any data augmentation methods in this study. Alternatively, the use of vision foundation models with lightweight fine-tuning can partially address the class imbalance issue, as these models often capture generalized features during pre-training, enabling them to better represent minority classes compared to conventional task-specific models trained solely on imbalanced datasets \citep{shi2024longtail}. Meanwhile, other strategies, including modifications to the loss function or the integration of multi-head architectures, can be explored and applied to further address the issue of class imbalance in future studies \citep{Rezaei-Dastjerdehei2020loss, Zhang2021multi-head}.

Although the model's transferability to multi-temporal and multi-spatial datasets was tested in this study, the sampling site was limited to Koh Tao. The adaptability of the model to broader regions outside the Indo-Pacific remains unexplored. Therefore, future research that uses coral reef images collected from other regions, such as the Red Sea, the Western Indian Ocean, the South China Sea, the Caribbean Sea, and the Gulf of Mexico, is essential to further evaluate the transferability of the proposed DINOv2-LoRA. To achieve this aim, we considered using images from public datasets such as SQUIDLE+ \citep{SquidlePlus2019} and CoralNet \citep{Beijbom2015CoralNet} as test sets. However, these images were captured with different cameras and at varying distances from the object, resulting in inconsistencies in image quality, color, and resolution, making it challenging even for domain experts to accurately label coral reef conditions \citep{Hunter2013publicdata}. In future work, direct collaborations with local conservation communities or research institutes could provide higher-quality images for further transferability testing.

Although the proposed DINOv2-LoRA demonstrated promising accuracy and efficiency in classifying coral reef conditions, several technical challenges remain unresolved before end users can directly use it. First, because coral reef ecosystems are highly sensitive to environmental changes, the characteristics of coral reef images may evolve in response to factors such as climate change \citep{Beyer2018climatechange}. For example, massive bleaching events caused by global warming can destroy entire habitats, leading to shifts to algae-dominated ecosystems \citep{Randazzo-Eisemann2021algae}. Such changes in the features captured in the images can confuse the model and lead to decreased accuracy. The model must be continuously updated with the latest coral reef images as part of its training set to address this. Future work could explore the integration of continual learning techniques such as Online Test-Time Adaptation (OTTA), allowing the model to update itself continuously without requiring offline retraining \citep{Boudiaf2022OTTA}. This approach would allow the model to adapt to multi-year datasets or data collected from various regions \citep{Wang2022otta}. Second, although we make the codes for establishing the model open-sourced, it remains challenging for end users without coding knowledge to apply the model in practice. Therefore, a user-friendly Graphical User Interface (GUI) could be developed to package the model, making it more accessible to non-professional stakeholders and conservation volunteers \citep{Pavoni2022GUI}.

\section{Conclusions}
While deep learning models are increasingly applied for automatically classifying coral reef conditions, conventional deep learning models face significant challenges in achieving high accuracy and transferability in this complex, multi-temporal, and multi-spatial underwater ecological task. In this study, we propose a novel method that integrates the vision foundation model DINOv2 with LoRA as an adapter for the multi-label classification of coral reef conditions. Using field images collected during dry and wet seasons at 15 diving sites in Koh Tao, Thailand, we compared and evaluated the performance and efficiency of various conventional deep learning models against our proposed DINOv2-LoRA. The experimental results demonstrate that DINOv2-LoRA achieved superior performance on the all-season mixup dataset while exhibiting exceptional generalization and adaptability to unseen seasons and habitats. This study pioneers a comprehensive assessment of the adaptation of foundation models for coral reef condition monitoring tasks under multi-temporal and multi-spatial settings. Our findings provide a benchmark for the development of models in this field and serve as a practical tool for the accurate and efficient monitoring of coral reef conditions in citizen science-based coral conservation programs. Future work could involve integrating continual learning techniques to enable adaptation to multi-year datasets or data collected from other regions. A user-friendly GUI could also be developed to package the model, making it accessible to non-professional stakeholders and conservation volunteers. Our proposed method can potentially help preserve and manage these valuable marine ecosystems.

\section*{Author's contributions}
\textbf{Xinlei Shao:} conceptualization (equal); data curation (lead); formal analysis (equal); funding acquisition (lead); investigation (lead); methodology (equal); project administration (lead); resources (supporting); software (supporting); writing - original draft (lead); writing – review and editing (equal). \textbf{Hongruixuan Chen:} conceptualization (equal); data curation (equal); formal analysis (equal); methodology (lead); software (lead); writing – original draft (equal); writing – review and editing (equal). \textbf{Fan Zhao:} investigation (equal); methodology (supporting); writing - review and editing (supporting). \textbf{Kirsty Magson:} data curation (supporting); investigation (supporting); resources (lead); writing – review and editing (equal). \textbf{Jundong Chen:} investigation (supporting); writing - review and editing (supporting). \textbf{Peiran Li:} methodology (supporting); software (supporting); writing - review and editing (supporting). \textbf{Jiaqi Wang:} investigation (supporting); writing - review and editing (supporting). \textbf{Jun Sasaki:} supervision (lead); writing - review and editing (lead). 

\section*{Acknowledgements}
This work was supported by the Sasakawa Scientific Research Grant from The Japan Science Society (Research No. 2023-2010), the Next Generation AI Research Center at The University of Tokyo, and the Graduate School of Frontier Sciences at The University of Tokyo, through the Challenging New Area Doctoral Research Grant (Project No. C2301). The authors thank the New Heaven Reef Conservation Program, Koh Tao, Thailand, for assisting with the field survey, data annotation, and marine biology and conservation diving training.

\section*{Data availability statement}
The coral image dataset, labels, and code are available at https://github.com/XL-SHAO/DINOv2-LoRA-coral-condition.

\bibliographystyle{apalike} % Set bibliography style to plainnat for alphabetical sorting

\end{document}